\documentclass{tlp}

\usepackage{amsmath}
\usepackage{times}
\usepackage{latexsym}

\newtheorem{definition}{Definition}[section]
\newtheorem{example}{Example}[section]

\newtheorem{lemma}{Lemma}[section]
\newtheorem{theorem}[lemma]{Theorem}
\newtheorem{corollary}[lemma]{Corollary}
\newtheorem{proposition}[lemma]{Proposition}
\newcommand{\qed}{\mbox{$\Box$}}

\begin{document}
\title[Characterizations of Stable Model Semantics for Constraint Programs]
{Characterizations of Stable Model Semantics for Logic Programs with Arbitrary Constraint Atoms}
\author[Y.-D. Shen, J.-H. You, L.-Y Yuan]
{YI-DONG SHEN \\
State Key Laboratory of Computer Science, Institute of Software,
Chinese Academy of Sciences, Beijing 100190, China\\
\email{ydshen@ios.ac.cn}
\and JIA-HUAI YOU, LI-YAN YUAN\\
Department of Computing Science, University of
Alberta, Edmonton, Alberta, Canada T6G 2H1\\
\email{\{you,yuan\}@cs.ualberta.ca}
}

\submitted{22 March 2008}
 \revised{1 Dec 2008, 29 March 2009}
 \accepted{28 April 2009}

\maketitle

\begin{abstract}
This paper studies the stable model semantics of 
logic programs with (abstract) constraint atoms and their
properties. We introduce 
a succinct abstract representation of these constraint atoms 
in which a constraint atom is 
represented compactly.
We show two applications.
First, under this representation of constraint atoms,
we generalize the Gelfond-Lifschitz 
transformation and apply it to define stable models (also called answer sets) 
for logic programs
with arbitrary constraint atoms. The resulting semantics turns out to 
coincide with the one defined by Son et al.,
which is based on a 
fixpoint approach.
One advantage of our approach is 
that it can be applied, 
in a natural way, to define stable models for 
disjunctive logic programs with constraint atoms, which 
may appear in the disjunctive head
as well as in the body of a rule. 
As a result, our approach to the stable model semantics 
for logic programs with constraint atoms
generalizes 
a number of previous approaches.
Second, we show that our abstract representation of constraint atoms provides
a means to characterize dependencies of atoms 
in a program with constraint atoms, so that some standard characterizations
and properties relying on these dependencies in the past for 
logic programs with ordinary atoms can be extended to 
logic programs with constraint atoms.
\end{abstract}
   
\section{Introduction}
\label{int}
Answer set programming (ASP) as an alternative logic programming paradigm
has been demonstrated to 
be an effective knowledge representation formalism for 
solving combinatorial search problems arising in many
application areas such as planning, reasoning about 
actions, diagnosis, abduction, and so on \cite{Baral03,GL02,Lif02,MT99,Nie99}.  
In recent years, researchers 
have paid particular attention to extensions of ASP 
with means to model aggregate constraints  
in particular, and constraints on sets in general 
\cite{CFLP05,DFILP03,DPB01,EPS04,EPS05,FLP04,Ferr05,LPST07,LT05,LT06,MNT06,MR04,MT04,Pel04,PDB03,PDB07,PT04,SY07,SNS02,SP06-1,SPT06,SPT07}.
Logic programs with constraint atoms were introduced
as a general framework for representing, and reasoning with, sets of atoms
\cite{MNT06,MR04,MT04}. This is in contrast 
with traditional logic programs, which are used primarily 
to reason with individuals.

The abstract form of a constraint atom takes the form
$(D,C)$, where $D$ is a finite set of atoms and $C$ a collection
of subsets from the power set of $D$, which  
expresses a constraint on the domain $D$ 
with the collection $C$ of admissible solutions. 

By allowing constraint atoms to appear anywhere in a rule, the framework of
logic programs has become a highly expressive
knowledge representation language. For example, 
many constraints can be conveniently and compactly represented with constraint
atoms 
such as weight
and cardinality constraints and aggregates 
(see, e.g. \cite{CFLP05,DFILP03,DPB01,FLP04,MT04,Pel04,SNS02}).
In fact, any constraint studied in the context of 
constraint satisfaction problem (CSP) can be represented by 
a constraint atom.
In this way, the framework of 
logic programs with constraint atoms can express 
complex constraint satisfaction 
problems, such as those involving conditional constraints
\cite{Cond-CSP-introd} (called {\em dynamic CSPs}),
which are useful in modeling configuration and design problems.
When the head of a rule is allowed to be a disjunction of constraint atoms, 
logic programs become capable of expressing, not only conditional
constraints, but also  
disjunctive constraints,
both of which have been investigated by 
the constraint programming community 
outside of logic programming (see, e.g.
\cite{disjunctive-constraint,disj-JACM,disjunctive-CSP}). 
For example, disjunctive constraints have been 
widely used in scheduling
to ensure that the time intervals over which activities
require the same resource do not overlap in time 
\cite{disjunctive-constraint}. Although practical ASP languages and 
systems typically incorporate concrete, predefined constraint atoms, such as 
weight constraint atoms \cite{SNS02} and aggregate atoms \cite{dell-faber-ijcai-03}, the adoption of the abstract form of constraint atoms has made it possible to study 
the semantics and properties of these programs in an abstract setting. 

In this paper, we characterize and define stable models 
for logic programs with constraint atoms by introducing 
a succinct abstract representation of constraint atoms. 
In the current literature as mentioned above, a constraint atom 
is expressed as a pair $(D,C)$,
where $D$ is a finite set of ground atoms and $C$ a collection
of sets of atoms in $D$. We call this a 
{\em power set
 form} representation (w.r.t. $D$)
of constraint atoms, as $C$ may involve the whole power set $2^D$ of $D$.
This is the case even for special classes of constraint atoms such as 
monotone constraint atoms (a constraint atom $(D,C)$ is {\em monotone} if 
for any $S\subset D$, whenever $S\in C$ all of its supersets in $2^D$ are 
in $C$). 

For instance, suppose we have a monotone constraint atom $A_1 = (D,2^D)$. Semantically,
this constraint atom is a tautology, since for any set $I$ of atoms, it is a fact
that $I$ satisfies $A_1$, in the sense that 
$I \cap D \in 2^D$.
A clever representation would just need to express two pieces of information,
the ``bottom element'' $\emptyset$ and the ``top element'' $D$; two elements 
together
implicitly cover all the sets in between.
So, instead of using the power set representation to express all the admissible
solutions of this constraint atom, we could use a pair of sets.
As another example, consider a monotone constraint atom
$A_2 = (D,2^D \setminus \{\emptyset\})$. A minimal element (set inclusive)
in 
$2^D \setminus \{\emptyset\}$ 
is a singleton in $2^D$. In this case, any minimal element $B$ in 
$2^D$ and $D$ form a pair with $B$ being the bottom element and $D$ being 
the top. So, we could represent this constraint atom by a collection of pairs,
one for each singleton in $D$. The number of such pairs in this case equals to
the size of $D$.

In this paper, we introduce such an abstract
representation.
In general, the abstract representation
of a constraint atom $(A_d, A_c)$ is expressed as
$(A_d,A_c^*)$, where 
$A_c^*$ consists of what will be called 
{\em abstract prefixed power sets}, denoted $W\uplus V$,
with $W, V\subseteq A_d$ and $W\cap V = \emptyset$.
Intuitively, $W\uplus V$ represents a collection 
of sets of the form $W \cup S$ with $S \in 2^V$, all of which are in $A_c$.

The abstract representation of constraint atoms not only yields
a compact representation, but also captures the 
essential information embodied in constraint atoms appearing in the bodies
of rules.
To substantiate this claim, we show two applications. 

In the first application, we restore the power of the Gelfond-Lifschitz 
transformation by generalizing it for logic programs with 
constraint atoms.
The key idea is that
given an interpretation $I$,
each constraint atom $A=(A_d,A_c^*)$ under our abstract representation
can be concisely characterized by a set of 
{\em abstract satisfiable sets} of the form $W\uplus V\in A_c^*$ 
such that $W\uplus V$ covers $I\cap A_d$. Therefore, the standard 
Gelfond-Lifschitz transformation can be naturally generalized to logic 
programs 
with constraint atoms by representing each constraint atom by its
abstract satisfiable sets. 
We then use the generalized Gelfond-Lifschitz transformation 
to define stable models for
disjunctive logic programs with constraint atoms. 
It turns out that, for non-disjunctive logic programs
with arbitrary constraint atoms,
the stable models 
defined this way are precisely those 
defined by Son et al. 
\cite{SPT06,SPT07} for logic programs with arbitrary constraint atoms, and the equivalent semantics, called 
the {\em ultimate stable semantics}, for 
aggregate logic programs
\cite{DPB01}. These semantics are defined by  
a substantially different approach, the 
fixpoint approach.

One advantage of our approach is that
the semantics is defined 
for
disjunctive programs where
a constraint atom can appear anywhere in a disjunctive rule.
This is due to the power of the Gelfond-Lifschitz transformation.
Roughly speaking, for a non-disjunctive program with constraint atoms, 
a stable model $M$
is just the least model of the reduct by the 
generalized Gelfond-Lifschitz transformation, while for 
a disjunctive program with constraint atoms, a stable model $M$
is a minimal model of the reduct. 
We show that logic programs whose constraint atoms appearing in disjunctive rule
heads
are elementary possess the minimality property; i.e., for such logic programs,
all stable models under the  
generalized Gelfond-Lifschitz transformation are minimal models. 
Thus, by the known relationships among different definitions of stable models,
the stable model semantics defined in this paper
extends the semantics of conventional disjunctive logic programs
\cite{GL91}, 
the semantics defined by Marek and Truszczynski \cite{MT04}
for non-disjunctive logic programs with monotone constraint atoms, the 
semantics by Son et al. \cite{SPT06,SPT07}, and others equivalent to it
\cite{DPB01,PDB03}. 

We note that disjunctive programs with aggregates
have 
been studied previously in \cite{FLP04,PT04},
where aggregates do not appear in the heads of program rules.

In the second application,
we show that our abstract representation of constraint atoms provides
a means to characterize the {\em dependency relation}
among ordinary atoms 
in a program with constraint atoms. This dependency relation in the past is
defined 
using
a {\em dependency graph}. One question for logic programs
with constraint atoms is how this dependency graph may be constructed so that
the means to characterize the properties of programs by {\em odd cycles}, 
{\em even cycles}, 
{\em call-consistency}, {\em acyclic programs} in the traditional context 
is applicable to the new context.
We show that the information captured in
our abstract representation
is essential in constructing the 
dependency graph for a program with constraint atoms. As we will see, this is due to 
a simple way to represent a logic program with constraint atoms by a normal logic 
program.

To summarize, the main contributions of this paper are:
\begin{itemize}
\item 
We introduce an abstract yet compact representation 
of constraint atoms, 
independent of any programs in which they appear. 
\item 
Using this abstract representation, we present a generalized form of 
Gelfond-Lifschitz transformation and apply it to define 
stable models for disjunctive logic programs with constraint atoms.
For non-disjunctive programs, the semantics defined this way 
coincides with the one 
based on conditional satisfaction 
\cite{SPT06,SPT07}, and with the ones equivalent to it \cite{DPB01}. 
\item 
We show that our abstract representation
of constraint atoms encodes the information needed to capture the atom dependency 
relation in a given program, thus the means to characterize 
the properties for normal programs
can still be applied to programs with constraint atoms, and in the process,
the unfolding approach \cite{SP06-1}
is made simple.
\end{itemize}

The paper is structured as follows. 
Following the preliminaries in the next section,
in Section 
\ref{sec-abs} we present our 
abstract representation of constraint atoms. In Section \ref{sec-chara-catom},
we show some characterization of constraint atoms
under this abstract representation.
In Section \ref{sec-genGL},
we introduce a generalized Gelfond-Lifschitz 
transformation and apply it to define stable models for
disjunctive logic programs with constraint atoms. 
In Section \ref{sec-Son}, we prove the relationship of our
approach with Son et al.'s fixpoint approach \cite{SPT06}.
Then in Section \ref{properties}, we show 
that our abstract representation of constraint atoms encodes 
precisely the needed information to define 
the dependency graph of a program with constraint atoms. 
In Section \ref{sec-rel}, we discuss the related approaches.
Finally in Section \ref{sec-concl}, we provide conclusions and discuss future work.

Proofs of theorems and lemmas will be delayed to Appendix. 

Some results of this paper have been reported in \cite{SY07}. 
The current paper, however, contains substantial new results. 

\section{Preliminaries}
\label{sec2}
We consider propositional (ground) logic programs and assume a fixed
propositional language with a countable set 
${\cal V}$ of propositional atoms (atoms for short).
Any subset $I$ of ${\cal V}$ is called an {\em interpretation}.
A literal is an atom $A$ (a {\em positive literal}) 
or its negation $not\ A$ (a {\em negative literal}).
For a set $S = \{A_1, ..., A_m\}$ of atoms,
we use $not\ S$ to denote $\{not\ A_1, ..., not\ A_m\}$
and $|S|$ to denote the size of $S$. 
For convenience, when $S$ appears in a logic expression,
it represents a conjunction $A_1\wedge ...\wedge A_m$;
when $not\ S$ appears in a logic expression,
it represents a conjunction $not\ A_1\wedge ...\wedge not\ A_m$.

An {\em abstract constraint atom} (or {\em c-atom} following \cite{SPT06,SPT07}) $A$ is a pair $(D,C)$,
where $D$ is a finite set of atoms in ${\cal V}$ and $C$ a collection
of sets of atoms in $D$, i.e., $C\subseteq 2^D$. For convenience,
we use $A_d$ and $A_c$ to refer to the components 
$D$ and $C$ of $A$, respectively. 
As a general framework, c-atoms can be used to represent
any constraints with a finite set $A_c$ of admissible solutions
over a finite domain $A_d$. 

A c-atom $A$ is {\em elementary} if it is of 
the form $(\{a\}, \{\{a\}\})$, where $a$ is an atom.
Due to the equivalence in satisfaction,
an elementary c-atom may be simply written by the atom in it.
$A$ is {\em monotone} if it has the property that
for any $S\subset A_d$, if $S\in A_c$ then all of 
its supersets in $2^{A_d}$ are in $A_c$. 
$A$ is {\em nonmonotone} if it is not monotone.
$A$ is {\em antimonotone} if $A_c$ is closed under subsets, i.e.,
for every $X,Y \subseteq A_d$, if $Y \in A_c$ and $X \subseteq Y$ then $X \in A_c$.
$A$ is {\em convex} if for any $S_1, S, S_2\subseteq A_d$
such that $S_1\subseteq S\subseteq S_2$ and $S_1,S_2\in A_c$,
we have $S\in A_c$.
  
A {\em disjunctive constraint program}, 
or a {\em disjunctive (logic) program 
with c-atoms},
is a finite set of rules of the form
\[H_1\vee ...\vee H_k \leftarrow A_1, ..., A_m, not\ B_1, ..., not\ B_n\]
where $k\geq 1$, $m,n \geq 0$ and $H_i$, $A_i$ and $B_i$ are either an atom or a c-atom 
(``$\leftarrow$" is omitted when $m=n=0$). 
$P$ is a {\em normal constraint program} if $k=1$ for all of its rules; 
$P$ is a {\em positive constraint program} if $n=0$ for all of its rules;
$P$ is a {\em positive basic program} if $n=0$ and 
$k=1$ with $H_1$ being an elementary c-atom for all of 
its rules. 
As usual, $P$ is a {\em normal program} if $P$ is a
normal constraint program where all c-atoms are elementary;
$P$ is a {\em disjunctive program} if $P$ is 
a disjunctive constraint program 
where all c-atoms are elementary. 

In the sequel, if not specifically quantified, 
a logic program
(or simply
a program) refers to a disjunctive constraint program. 
To make it explicit, when a program contains only 
elementary c-atoms, it may be called a program with 
{\em ordinary} atoms, or
just a program without c-atoms. 

Let $r$ be a rule of the above form. We define 
\begin{tabbing}
\hspace{.2in} 
$head(r) = \{H_1,...,H_k\}$\\
\hspace{.2in} 
$body(r) = \{A_1, ..., A_m, not\ B_1, ..., not\ B_n\}$
\end{tabbing}
which will be referred to as 
the {\em head} and the {\em body} of the rule, respectively,
where $body(r)$ denotes the conjunction of the elements in the set
and $head(r)$ the disjunction.
Without confusion, we may use the set notation in a rule to express the body
as well as the head. For example, given a rule $r$, we may write 
$head(r) \leftarrow body(r)$. 

We will use $At(P)$ to denote the set of atoms that appear in a program $P$.

The {\em satisfaction} relation is defined as follows.
An interpretation $I\subseteq {\cal V}$ satisfies an atom $a$ if $a\in I$;
$not\ a$ if $a\not\in I$. 
$I$ satisfies a c-atom $A$ if $I \cap A_d\in A_c$;
$not\ A$ if $I \cap A_d\not\in A_c$.
This relation extends
to arbitrary expressions $F$ mentioning negation $not$, conjunction $\wedge$
and disjunction $\vee$, in a usual way.
We will use $I\models F$ to denote that $I$ satisfies $F$,
and $I\not\models F$ to denote that $I$ does not satisfy $F$.
We say $F$ is {\em true} (resp. {\em false}) in $I$ if and only if
$I$ satisfies (resp. does not satisfy) $F$. 

$I$ satisfies a rule $r$ if it satisfies $head(r)$
or it does not satisfy $body(r)$. $I$ is a {\em model} 
of a logic program $P$ if 
it satisfies all rules of $P$. $I$ is a {\em minimal model} of $P$ if it is a model
of $P$ and there is no proper subset of $I$ which is also a model of $P$.
$I$ is a {\em supported} model of $P$ if for any $a\in I$, 
there is $r \in P$ such that $a \in head(r)$ and $I\models body(r)$. 

As commented earlier, atoms can be viewed as elementary c-atoms. This is due to 
the fact that 
for an atom $a$, an interpretation $I$ satisfies $a$ iff $a \in I$ iff
$I \models (\{a\},\{\{a\}\})$. 

Sometimes we say {\em a model M restricted to the atoms appearing in a 
program $P$}. By this we mean $M \cap At(P)$, and denote it by 
$M|_{At(P)}$.

Note that c-atoms of the form $(D,\emptyset)$ are not satisfied by any
interpretation. We will use a special symbol $\bot$ to denote any such 
c-atom. 

Following \cite{SPT07},
for any c-atom $A = (A_d, A_c)$, its negation $not\ A$
is interpreted by its complement, which is 
a c-atom $(A_d, A_c^-)$ where $A_c^-
= 2^{A_d} \setminus  A_c$.\footnote{Note that this is consistent with our definition of satisfaction of negated c-atoms.  
But not all semantics are based on the complement 
approach. A detailed comparison can be found in \cite{SPT07}.}
So a logic program with negated c-atoms can
be rewritten to a logic program free of negated c-atoms by replacing all 
occurrences of negated c-atoms with their respective complement c-atoms.
Due to this assumption, in the sequel we only consider logic programs without
negated c-atoms in rule bodies.

Given a disjunctive program $P$ (where c-atoms are elementary) and an interpretation $I$,
the standard {\em Gelfond-Lifschitz transformation} of $P$ w.r.t.$I$, written as $P^I$,
is obtained from $P$ by performing two operations: 
(1) remove from $P$ all rules whose bodies contain a negative 
literal $not\ A$ such that $I\not\models not\ A$, and (2)
remove from the remaining rules all negative literals.
Since $P^I$ is a positive constraint program where c-atoms are elementary,
it has a set of minimal models. 
$I$ is defined to be a stable model of $P$ if it is a minimal model
of $P^I$ \cite{GL88,GL91,Przy91}.

The cardinality and weight constraints can be represented
by c-atoms. In some of the example programs of this paper 
we may write weight constraints instead of c-atoms. 
We will adopt the notation
proposed in \cite{SNS02}. A {\em weight constraint} is an expression of the form
$$l \{a_1 = w_{a_1}, ..., a_n = w_{a_n}, not  \ b_1 =w_{b_1}, ..., not \ b_m =
w_{b_m}\}u$$
where $a_i$ and $b_j$ are atoms and $w_{a_i}$ is the weight of atom $a_i$ and 
$w_{b_j}$ is the weight of negative literal $not \ b_j$. The numbers $l$ and $u$ are lower and upper bounds of the constraint, respectively. A weight constraint is satisfied 
by a set of atoms $S$ if the sum of the weights of the literals in the set
$\{a_1 , ..., a_n, not\  b_1,..., not\ b_m\}$ that are satisfied by $S$ is between $l$ and $u$ (inclusive).

A {\em cardinality constraint}
is a special case of weight constraint where
each weight is one. In writing a cardinality constraint, we will omit 
the weights. A {\em choice constraint} is a 
cardinality constraint of the form
$l \{a_1,..., a_n\} u$,
where $l =0$ and $u =n $. 
In writing a choice constraint, we will omit the bounds. 

\section{Abstract Representation of Constraint Atoms}
\label{sec-abs}
In this section, we present a compact 
representation of c-atoms. In the current literature,
for any c-atom $A$ its admissible solutions are all
explicitly enumerated and written in $A_c$. In many cases,
$A_c$ may involve a large portion of 
$A_d$.
It is then of
great interest if we can represent $A_c$
using some abstract structure so that its size can be substantially compressed.
We begin by introducing a notion of prefixed power sets.

\begin{definition}
Let $I=\{a_1,...,a_m\}$ and $J=\{b_1,...,b_n\}$  $(m,n\geq 0)$
be two sets of atoms. 
\begin{enumerate}
\item
The {\em $I$-prefixed power set} of $J$, denoted by $I\uplus J$,
is the collection $\{I\cup J_{sub} |J_{sub}\in 2^J\}$; i.e., each set in the collection 
consists of all $a_i$s in $I$ plus zero or more $b_i$s in $J$.
For any set of atoms $S$, we say 
$S$ is {\em covered by} $I\uplus J$ (or $I\uplus J$ {\em covers} $S$)
if $I\subseteq S$ and $S \subseteq I\cup J$.
\item
For any two abstract prefixed power sets $I\uplus J$ and $I'\uplus J'$,   
$I\uplus J$ {\em is included} in $I'\uplus J'$ 
if any set covered by $I\uplus J$ is covered by $I'\uplus J'$. 
\end{enumerate}
\end{definition}

\begin{theorem}
\label{t1}
When $I\uplus J$ is included in $I_1\uplus J_1$, 
we have $I_1\subseteq I$ and 
$I\cup J\subseteq I_1\cup J_1$. 
If $I\uplus J$ is included in $I_1\uplus J_1$
and $I_1\uplus J_1$ is included in 
$I_2\uplus J_2$, then $I\uplus J$ is included in 
$I_2\uplus J_2$.
\end{theorem}

Given a c-atom $A$, let $I \in A_c$ and $J\subseteq A_d \setminus I$.
$I\uplus J$ is called {\em I-maximal in $A$} (or just {\em maximal})
if all sets covered by $I\uplus J$ are in $A_c$ and  
there is no $J'$ with $J \subset J' \subseteq A_d \setminus I$ such that
all sets covered by $I\uplus J'$ are in $A_c$ .
 
\begin{definition}
\label{ab-1}
Let $A$ be a c-atom and $S\in A_c$. 
The collection of {\em abstract $S$-prefixed power sets} 
of $A$ is 
$\{S\uplus J \mid S\uplus J \mbox{ is } S\mbox{-maximal in } A\}$.
\end{definition}

For instance, consider a c-atom $A$, where
$$
\begin{array}{ll}
A_d = \{a,b,c,d\} \\
A_c = \{\emptyset, \{b\},\{c\}, \{a,c\}, \{b,c\},\{c,d\},\{a,b,c\}, \{b,c,d\}\}.
\end{array}
$$
For $\emptyset \in A_c$, the collection of abstract $\emptyset$-prefixed power sets 
of $A$ is $\{\emptyset\uplus \{b,c\}\}$; for $\{b\} \in A_c$, the collection 
is $\{\{b\}\uplus \{c\}\}$; for $\{c\} \in A_c$, the collection 
is $\{\{c\}\uplus \{a,b\},\{c\}\uplus \{b,d\}\}$. 
Note that $\{b\}\uplus \{c\}$ is included in $\emptyset\uplus \{b,c\}$. 
It is easy to check that all abstract prefixed power sets
for $\{a,c\}, \{b,c\}, \{a,b,c\}\in A_c$ are included in $\{c\}\uplus \{a,b\}$ 
and all those for $\{b,c\}, \{c,d\}, \{b,c,d\}\in A_c$
are included in $\{c\}\uplus \{b,d\}$. 

When a collection contains two abstract prefixed power sets,
$I\uplus J$ and $I'\uplus J'$ with $I\uplus J$ being included in $I'\uplus J'$,
we say $I\uplus J$ is {\em redundant} in this collection. 

For instance,
consider $I\uplus J$ where 
$I = \{a,b\}$ and $J = \{c\}$, and 
$I'\uplus J'$ where 
$I' = \{a\}$ and $J' = \{b, c\}$. Then,  
$I\uplus J$ is redundant in a collection that contains 
$I'\uplus J'$, since every set covered by 
$I\uplus J$ is covered by $I'\uplus J'$.

\begin{definition}
\label{ab-2}
The {\em abstract representation} $A^*$ of a c-atom $A$ is a pair $(A_d, A_c^*)$
where $A_c^*$ is the collection $\bigcup_{S\in A_c} C_S$, where $C_S$ is the
collection of abstract $S$-prefixed power sets of $A$, with
all redundant abstract prefixed power sets removed.
\end{definition}

Observe that when $W\uplus V$ is in $A_c^*$, all sets in the
collection $\{W\cup V_{sub} |V_{sub}\in 2^V\}$ are in $A_c$.
Conversely, when $\{W\cup V_{sub} |V_{sub}\in 2^V\}\subseteq A_c$,
there exist $W',V' \subseteq A_d$ such that 
$W' \subseteq W$ and $W \cup V \subseteq  W' \cup V'$, and  
$W' \uplus V' \in A^*_c$, i.e.,
$W\uplus V$ is included in $W'\uplus V'\in A_c^*$.
In other words,
$A^*_c$ is the collection of 
maximal sublattices of the lattice $(2^{A_d},\subseteq)$, of which 
all elements are in $A_c$. For such a maximal sublattice 
$W \uplus V$,
the bottom element is 
$W$ and the top element is $W \cup V$.

Consider the above example c-atom $A$ again.
Its abstract representation is $(A_d, A_c^*)$ 
with $A_c^* = \{\emptyset\uplus \{b,c\}, \{c\}\uplus \{a,b\},\{c\}\uplus \{b,d\}\}$.

\begin{theorem}
\label{ab-form}
Let $A=(A_d,A_c)$ be a c-atom. 
\begin{enumerate}
\item[{\rm (1)}]
$A$ has a unique abstract form $(A_d,A_c^*)$.
\item[{\rm (2)}]
For any interpretation $I$, $I\models A$ if and only if $A_c^*$ contains an 
abstract prefixed power set $W\uplus V$ covering $I\cap A_d$. 
\end{enumerate}
\end{theorem}

For some special classes of c-atoms, their abstract representations are 
much simpler and can be stated more structurally.

We need a terminology:
given a set $S$ 
of sets, we say that 
$I \in S$ is {\em minimal} (resp. {\em maximal}) in $S$ if 
there is no $I' \in S$ such that $I' \subset I$ (resp. $I' \supset I$).

%The following result follows immediately from 
%the definitions of monotone, antimonotone, and convex c-atoms.

\begin{theorem}
\label{mono-conv-catom}
Let $A$ be a c-atom. 
\begin{enumerate}
\item[{\rm (1)}]
$A$ is monotone if and only if 
$A_c^* =\{B \uplus A_d \setminus B :  \ B \mbox{ is minimal in } A_c\}$
if and only if 
$|W| + |V| = |A_d|$
for each $W\uplus V\in A_c^*$. 
 
\item[{\rm (2)}] 
$A$ is antimonotone if and only if 
$A_c^* =\{\emptyset \uplus T  :  \ T \mbox{ is maximal in } A_c\}$
if and only if 
$W = \emptyset$ 
for each $W\uplus V\in A_c^*$. 

\item[{\rm (3)}] 
$A$ is convex if and only if 
$A_c^* =\{B \uplus T  : \ 
B  \mbox{ is minimal and } B\cup T \mbox{ is maximal in } A_c\}$.  
\end{enumerate}
\end{theorem}

By this theorem, given $A^*$, to check if $A$ is monotone (resp. antimonotone) it suffices to
check if $|W| + |V| = |A_d|$ (resp. $W = \emptyset$) 
for each $W\uplus V\in A_c^*$. This process 
takes linear time in the size of $A_c^*$. 
Let $S_1 = \{W \mid W\uplus V\in A_c^*\}$ and 
$S_2 = \{W\cup V \mid W\uplus V\in A_c^*\}$. 
To check if $A$ is convex,
it suffices to check (i) there are no $p,q\in S_1$ with $p\subset q$,
and (ii) there are no $p,q\in S_2$ with $p\subset q$. 
Case (i) guarantees that 
$W$ is minimal while case (ii) guarantees $W\cup V$ is maximal in $A_c$,
for each $W\uplus V\in A_c^*$.
The time for the two cases is bounded by
$O(|A_c^*|^2 * |A_d|^2)$, where 
each subset check is assumed to take time $|A_d|^2$. 
This leads to the following complexity result.

\begin{theorem}
\label{mono-conv-comp}
Given the abstract representation $A^*$ of a c-atom $A$, 
the time to check if $A$ is monotone or antimonotone is linear in the size of $A_c^*$,
while the time to check if $A$ is convex is bounded by $O(|A_c^*|^2 * |A_d|^2)$.
\end{theorem}

We now discuss the issue of compactness. 
Given a c-atom $A$, its abstract representation $A^*$ is 
more compact than 
$A$ when $A_c$ contains some relatively large 
abstract prefixed power sets. 
This can be seen from the special classes of c-atoms in Theorem \ref{mono-conv-catom}.
Since the admissible solutions in such 
a c-atom are tightly clustered together,  they easily form large 
abstract prefixed power sets. 
For example, 
since a monotone c-atom is closed under its supersets in $A_c$, for 
any minimal set $S$ in $A_c$, all  
the sets in between $S$ and $A_d$ must be in $A_c$. Therefore, 
$S \uplus A_d \setminus S$ is an abstract $S$-prefixed power set.
The bigger is the difference between 
$S$ and $A_d$, the more information is captured compactly.
As another example, we know  that weight
constraints without negative literals or negative weights are convex.
That is, these 
constraints are of the  
form 
$l \{a_1 = w_{a_1}, ..., a_n = w_{a_n}\} u$,
where $w_{a_i} \geq 0$, for all $1\leq i \leq n$.
Let $A$ denote such a weight constraint. Then, $A_d = 
\{a_1,...,a_n\}$ and $A_c$ consists of all subsets of $A_d$ where 
the sum of the weights of the atoms in such a subset is between $l$ and $u$.
Thus, if the sets $B$ and $T$ are such that
$B \subseteq T \subseteq A_d$, and
$B$ is 
minimal and $T$ is maximal in $A_c$,
then $B \uplus T\setminus B$ forms
an abstract $B$-prefixed power set, representing all the sets in between.

Apparently, c-atoms that are {\em nearly} monotone 
(or antimonotone or convex) can greatly benefit from the abstract 
representation. For example, given a set $S = \{a_1,...,a_n\}$,
a c-atom that expresses 
all subsets of $S$ except some $V$ in between $\emptyset$ and $S$ can easily
fall outside of the above special classes. For instance, suppose $S =\{a,b,c\}$
and 
let $A =(S, 2^S \setminus  \{\{a,b\}\})$. Then $A^* = (S, \{\emptyset \uplus \{a,c\},
\emptyset \uplus \{b,c\}, \{a,c\} \uplus \{b\},
\{b,c\} \uplus \{a\}\})$.

It should also be clear that there are situations where $A^*$ may not be 
strictly more compact than $A$. This is typically the case where the admissible
solutions in $A_c$ are largely unrelated.
We say that 
two sets $I$ and $J$ are 
{\em unrelated} if either no one is a 
subset of the other, or 
$I \subseteq J$ and $J\setminus I$ is not singleton.

For example, consider a c-atom $A$ where $A_c$ consists of all subsets of 
$A_d$ with an equal size.
In this case, no set in $A_c$ is a subset of  
another in $A_c$. The abstract representation of such a c-atom $A$ 
is $(A_d,A^*_c)$ where
$A^*_c = \{I \uplus \emptyset : I \in A_c\}$,  which trivially enumerates 
all admissible solutions in $A_c$.
As another example,
consider 
a c-atom
$A = (\{a,b,c,d\}, \{\emptyset, \{a,b\}, \{a,b,c,d\}\})$. 
In this case, for any $I, J \in A_c$,
if $J$ is a superset of $I$, then $J\setminus I$ is not singleton. 
The abstract representation of $A$ is $(A_d, A^*_c)$, where
$A^*_c = \{\emptyset \uplus \emptyset, 
\{a,b\} \uplus \emptyset,
\{a,b,c,d\} \uplus \emptyset\}$. Again, $A^*_c$ essentially enlists 
all admissible solutions in 
 $A_c$.

Although all the evidence indicates that 
for any c-atom $A$ the number of abstract prefixed power sets in
$A^*_c$ is less than or equal to the number of admissible solutions in $A_c$, 
i.e. $|A^*_c| \leq |A_c|$, a rigorous proof for this claim seems challenging.
We leave this proof as an interesting open problem.

Finally in this section, we comment that for a c-atom $A$, 
it takes polynomial time in the size of $A$ to
construct $A^*$. This result will be useful in determining the complexity
of the semantics defined by 
the generalized Gelfond-Lifschitz transformation later in this paper.

Below, we give a bound for the construction. 
\begin{theorem}
\label{construction-abstract-rep}
Let $A$ be a c-atom. 
The time to construct $A^*$ from $A$  
is bounded by $O(|A_c|^4 * |A_d|^2)$.
\end{theorem}

\section{Characterizations of C-Atoms under Abstract Representation}
\label{sec-chara-catom}
In this section, we present some
characterizations of c-atoms under the abstract representation. 
Essentially, these characterizations are related to the 
fact that a c-atom can be semantically represented by a propositional formula. 

Recall that the standard semantics of a c-atom $A$ is defined by its 
satisfaction:
for any set of atoms $M$, $M \models A$ if and only if $M \cap A_d \in A_c$.
For nonmonotone c-atoms, a difficulty with this interpretation of the meaning
of a c-atom is that 
the iterative construction 
by the {\em one-step provability operator} \cite{LT06,MNT06}
may lead to 
an undesirable situation \--  there is no guarantee that once a c-atom is satisfied 
by a set of atoms $I$, it remains to be satisfied by an extension of $I$.

However, by definition, a set of atoms $M$ satisfies a c-atom $A$ if and 
only if $M$ satisfies the propositional formula that corresponds to 
the admissible solutions in $A_c$. This formula is 
a disjunction of conjunctions, each of which 
represents 
an admissible solution in $A_c$. 
As a propositional formula, it can be simplified to a logically equivalent one.
It turns out that this simplification process is significant as 
it reveals the nature of the information encoded in our 
abstract representation. Therefore, the main result of this section is to show
that the abstract representation of 
a c-atom encodes the ``simplest'' propositional formula, in the form of 
a disjunctive normal form (DNF).
We then use this insight to 
define what are called {\em abstract satisfiable sets},
which make it possible to define a new form of 
Gelfond-Lifschitz transformation. 

Below, we make it precise as what the formula is, and state
some facts which easily follow from the definition.

\begin{proposition}
\label{th-sem}
Let $A=(A_d,A_c)$ be a c-atom with $A_c = \{S_1, ..., S_m\}$, and  
$I$ be an interpretation.
The DNF
$C_1\vee ...\vee C_m$ for $A$
is defined as:
each $C_i$ is a conjunction $S_i \wedge not\ (A_d \setminus  S_i)$. 

\begin{itemize}
\item[{\rm (1)}] 
$I$ satisfies $A$ if and only if
$C_1\vee ...\vee C_m$ is true in $I$. 
\item [{\rm (2)}] 
$I$ satisfies $not\ A$ if and only if 
$not\ (C_1\vee ...\vee C_m)$ is true in $I$.
\end{itemize}
\end{proposition}

Given a c-atom $A$, the DNF $C_1\vee ...\vee C_m$ for $A$
can be simplified. In propositional logic, we have 
$(S \wedge \neg F) \vee (S \wedge F) \equiv S$, for any formulas $S$ and $F$. 

\begin{example}
Consider a monotone c-atom 
$A = (\{a,b\}, \{\{a\}, \{b\}, \{a,b\}\}).$
Its corresponding DNF is 
$(a\wedge not\ b) \vee (b\wedge not\ a) \vee (a\wedge b)$, 
which can be simplified 
as follows:
\begin{tabbing}
\hspace{.3in} \= $(a\wedge not\ b) \vee (b\wedge not\ a) \vee (a\wedge b)$ \\
\> $\equiv \underline{(a\wedge not\ b)\vee (a\wedge b)} \ \vee\  \underline{(b\wedge not\ a) \vee (a\wedge b)}$ \\
\> $\equiv a \vee b$ 
\end{tabbing}
Note that in the second line above
a disjunct in the previous DNF is added.
\label{simplify}
\end{example}

What is interesting is that the resulting propositional formula corresponds to
the abstract representation of $A$,
where $A^*_c = \{\{a\} \uplus \{b\}, \{b\} \uplus \{a\}\}$.  This 
correspondence is made precise in the following theorem.

\begin{theorem}
\label{sem-catom-2}
Let $A$ be a c-atom and $M$ be a set of atoms. $M \models A$ if and only if $M$ satisfies
\begin{equation}
\label{sem2}
\bigvee_{W\uplus V \in A_c^*} W \wedge not\ (A_d \setminus  (W\cup V))
\end{equation}
\end{theorem}

The proof of this theorem requires the following lemma.
\begin{lemma}
\label{catom-lem}
Let $E = \{a_1,...,a_m\}$ be a set of atoms and $F$
be a DNF covering
all possible interpretations on the $a_i$s, i.e.
\[F = \bigvee_{1\leq i\leq m, \ L_i\in \{a_i, not\ a_i\}} L_1\wedge ...\wedge L_m\] 
$F$ can be simplified to $true$ in propositional logic
by applying the following rule:
\begin{equation}
\label{s-rule}
\mbox{For any } S_1 \mbox{ and } S_2, 
(S_1\wedge L\wedge S_2)\vee (S_1\wedge not\ L\wedge S_2) \equiv S_1\wedge S_2
\end{equation}
\end{lemma}

Note that rule (\ref{s-rule}) is like the {\em resolution rule} 
in its underlying pattern, 
but it applies to a DNF while resolution applies to 
CNFs. 

Theorem \ref{sem-catom-2}
shows that the satisfaction of a c-atom $A$
can be simplified to (\ref{sem2}) in terms of its abstract 
representation by applying rule (\ref{s-rule}).

As a slightly more involved example, 
consider a c-atom 
$$B = (\{a,b,c,d\}, \{\{d\}, \{a\}, \{a,b\},\{a,c\},\{a,b,c\}\}).$$
The DNF for this c-atom is:
$$
\begin{array}{ll}
(d \wedge not\ a \wedge not\ b\wedge not\ c) \vee 
(a \wedge not\ b \wedge not\ c \wedge not\ d) \vee \\
(a \wedge b\wedge not\ c \wedge not\ d) \vee 
(a \wedge c\wedge not\ b \wedge not\ d) \vee (a\wedge b\wedge c \wedge not\ d).
\end{array}
$$
which can be simplified to
\[(d \wedge not\ a \wedge not\ b\wedge not\ c) \vee (a \wedge not\ d)\] 
each disjunct of which corresponds to a prefixed power set in
the abstract representation of $B$, i.e.,
$B^*_c = \{\{d\} \uplus \emptyset, \{a\} \uplus \{b,c\}\}$.

We say that a DNF is {\em maximally simplified} if it 
cannot be further simplified by applying rule (\ref{s-rule}). 

The following theorem shows that (\ref{sem2}) is maximally simplified.

\begin{theorem}
\label{sem-catom-3}
The semantic characterization (\ref{sem2}) of a c-atom $A$
is maximally simplified.
\end{theorem}

Theorems \ref{sem-catom-2} and \ref{sem-catom-3} suggest that
the satisfaction of c-atom $A$ can be described by its simplest DNF
given in (\ref{sem2}),
independently of any interpretations.
When we generalize the standard Gelfond-Lifschitz transformation 
for constraint programs, we can
apply a given interpretation to further simplify this DNF.
In the following, and in the rest of the paper, given an interpretation $I$,
for any c-atom $A$ we use $T_A^I$ to denote $I\cap A_d$
and $F_A^I$ to denote $A_d\setminus T_A^I $. 

We are ready to define {\em abstract satisfiable sets}.
\begin{definition}
\label{sat-sets}
Let $A$ be a c-atom and $I$ an interpretation.
$W\uplus V\in A_c^*$ is an {\em abstract satisfiable set} of $A$ w.r.t. $I$ 
if $W\uplus V$ covers $T_A^I$. In this case, $W$ is called
a {\em satisfiable set} of $A$ w.r.t. $T_A^I$.
We use $A_s^I$ to denote the set of abstract satisfiable sets of $A$ w.r.t. $I$. 
\end{definition}

The next two theorems characterize some properties of 
abstract satisfiable sets as well as satisfiable sets.
\begin{theorem}
\label{catom-chara}
Let $A$ be a c-atom and $I$ an interpretation. 
$I\models A$ if and only if $I \models
\bigvee_{W\uplus V\in A_s^I} W \wedge not\ (A_d \setminus  (W\cup V))$.
\end{theorem}

\begin{theorem}
\label{th-num-min}
Let $A$ be a c-atom and $I$ an interpretation. If $S$ is a satisfiable set of 
$A$
w.r.t. $T_A^I$,
then for every $S'$ with $S\subseteq S'\subseteq T_A^I$, we have $S'\in A_c$. 
\end{theorem}

\section{A Generalization of the Gelfond-Lifschitz Transformation}
\label{sec-genGL}
In this section we show that
the characterizations of c-atoms presented in the last section
can be used to generalize  
the standard 
Gelfond-Lifschitz transformation for logic programs 
with c-atoms. 

In the following, special atoms of the forms $\theta_A$, $\beta_A$ 
and $\bot$ will be used, where $A$ is a c-atom. 
Unless otherwise stated, we assume that these special atoms
will not occur in any given logic programs or interpretations.
Let $\Gamma_\theta$ and $\Gamma_\beta$ 
be the sets of special atoms prefixed with $\theta$ and $\beta$, respectively.
Let $\Gamma = \Gamma_\theta \cup \Gamma_\beta$.

\begin{definition}
\label{gen-smodel}
Given a logic program $P$ and an interpretation $I$,
the {\em generalized Gelfond-Lifschitz transformation} 
of $P$ w.r.t. $I$, written as $P^I$,
is obtained from $P$ by performing the following four operations: 
\begin{enumerate}
\item
Remove from $P$ all rules whose bodies contain either a negative literal $not\ A$
such that $I\not\models not\ A$ or a c-atom $A$ such that $I\not\models A$.
\item
Remove from the remaining rules all negative literals.
\item
Replace each c-atom $A$ in the body of a rule 
with a special atom $\theta_A$ and introduce a new rule
$\theta_A\leftarrow A_1, ..., A_m$ for each 
satisfiable set $\{A_1, ..., A_m\}$ of $A$ w.r.t. $T_A^I$.
\item
Replace each c-atom $A$ in the head of a rule with $\bot$ if $I\not \models A$, 
or replace it with a special atom $\beta_A$ and introduce a new rule
$B\leftarrow \beta_A$ for each $B\in T_A^I$, a new rule
$\bot \leftarrow B, \beta_A$ for each $B\in F_A^I$, and
a new rule $\beta_A \leftarrow T_A^I$. 
\end{enumerate}
\end{definition}

In the first operation, we remove all rules whose bodies are not satisfied in $I$
because of the presence of a negative literal or a c-atom that
is not satisfied in $I$. In the second operation, we remove all negative literals 
because they are satisfied in $I$. The last two operations transform
c-atoms in the body and head of each rule, respectively.

Each c-atom $A$ in the body of a rule is substituted by a special atom $\theta_A$.
By Theorem \ref{catom-chara}, $\theta_A$ can be defined by 
introducing a new rule $\theta_A\leftarrow W \wedge not\ (A_d \setminus  (W\cup V))$  
for each abstract satisfiable set $W\uplus V$. Since the negative part
$not\ (A_d \setminus  (W\cup V))$ is true in $I$, it can be removed from
the rule body following the standard Gelfond-Lifschitz transformation.
Note that the remaining part $W$ is a satisfiable set.
Therefore, in the third operation, $\theta_A$ is defined by
introducing a new rule $\theta_A\leftarrow A_1, ..., A_m$ for each 
satisfiable set $\{A_1, ..., A_m\}$ of $A$ w.r.t. $T_A^I$. 
  
When $I\models A$, each c-atom $A$ in the head of a rule
is replaced by a special atom $\beta_A$. 
Note that $\beta_A$ represents a conclusion  
that every $B\in T_A^I$ is true and every $B\in F_A^I$ is false in $I$.  
Such a conclusion is formulated, in the fourth operation, 
by introducing a new rule 
$B\leftarrow \beta_A$ for each $B\in T_A^I$, a new rule
$\bot \leftarrow B, \beta_A$ for each $B\in F_A^I$, and
a new rule $\beta_A \leftarrow T_A^I$. 
$\bot$ is a special atom meaning $false$.
The last rule comes from the rule $\beta_A \leftarrow T_A^I \wedge not\ F_A^I$,
where the negative part $not\ F_A^I$ is true in $I$ and thus is removed 
following the standard Gelfond-Lifschitz transformation.  
When $I\not \models A$, we replace $A$ with $\bot$.
In the case that $\bot$ appears in a disjunction 
$B_1\vee...\vee \bot \vee ... \vee B_m$ with $m>0$,
$\bot$ can be removed, as the satisfaction of 
the disjunction is determined by the $B_i$s.

Apparently, the generalized Gelfond-Lifschitz transformation coincides with the 
standard Gelfond-Lifschitz transformation when $P$ contains no c-atoms.

Since the generalized transformation $P^I$ is a positive logic program
without c-atoms,
it has minimal models. We then define the stable model semantics
of a constraint program in the same way as that of a logic program
with ordinary atoms. 

\begin{definition}
\label{my-stablemodel}
For any logic program $P$, an interpretation $I$
is a {\em stable model} of $P$ if $I = M\setminus  \Gamma$, where
$M$ is a minimal model of the generalized Gelfond-Lifschitz transformation $P^I$.
\end{definition}

Immediately, if $P$ is a normal constraint program, 
then $I$ is a stable model 
of $P$ if 
$I = M\setminus  \Gamma$ and 
$M$ is the least model of 
the generalized Gelfond-Lifschitz transformation $P^I$.
In other words, the extension to disjunctive constraint programs from 
normal constraint programs follows the same way as 
the extension to disjunctive programs from 
normal programs. 

Again, stable models of $P$ under the generalized Gelfond-Lifschitz transformation
coincide with stable models under the standard Gelfond-Lifschitz transformation
when $P$ has no c-atoms. In the following, unless otherwise stated, by 
stable models we refer to stable models 
under the generalized Gelfond-Lifschitz transformation.

\begin{example}
\label{prog-2}
Consider the following program:
\begin{tabbing} 
\hspace{.2in} $P_1:\quad$ \= $p(1)$. \\ 
\> $p(-1)\leftarrow p(2).$  \\ 
\> $p(2) \leftarrow \mbox{SUM}(\{X|p(X)\})\geq 1.$ 

\end{tabbing}
The aggregate constraint $\mbox{SUM}(\{X|p(X)\})\geq 1$ 
can be represented by a c-atom $A$ where
$$
\begin{array}{ll}
A_d = \{p(-1),p(1),p(2)\}, \\
A_c = \{\{p(1)\}, \{p(2)\}, \{p(-1),p(2)\},\{p(1),p(2)\}, \{p(-1),p(1),$ $p(2)\}\}.
\end{array}
$$
Its abstract representation is $(A_d, A_c^*)$ 
with 
$$A_c^* = \{\{p(1)\}\uplus\{p(2)\}, \{p(2)\}\uplus\{p(-1),p(1)\}\}.$$ 
Let us check if $I = \{p(-1), p(1), p(2)\}$ is a stable model of $P_1$
using the generalized Gelfond-Lifschitz transformation. 
The first two operations do not apply.
Since $I\models A$ with $T_A^I = I\cap A_d = \{p(-1), p(1), p(2)\}$, 
$A$ has only one abstract satisfiable set $\{p(2)\}\uplus\{p(-1),p(1)\}$, 
and thus it has only one satisfiable set $\{p(2)\}$
w.r.t. $T_A^I$. So, in the third operation $A$ is replaced
by a special atom $\theta_A$, followed by a new rule
$\theta_A \leftarrow p(2)$. Hence we have 
\begin{tabbing} 
\hspace{.2in} $P_1^I:\quad$ \= $p(1)$. \\ 
\> $p(-1)\leftarrow p(2).$  \\ 
\> $p(2) \leftarrow \theta_A.$ \\
\> $\theta_A \leftarrow p(2)$. 
\end{tabbing}
The only minimal model of $P_1^I$ is $\{p(1)\}$, 
so $I$ is not a stable model of $P_1$.

It is easy to check that this program has no stable model.
\end{example}

\begin{example}
\label{prog-4}
Consider a disjunctive constraint program:
\begin{tabbing} 
\hspace{.2in} $P_2:\quad$ \= $A\vee B$.\\
\> $a \leftarrow b.$ 
\end{tabbing}
where $A$ is a c-atom $(\{a\}, \{\{a\}\uplus \emptyset\})$
and $B =  (\{b\}, \{\{b\}\uplus \emptyset\})$.
\begin{enumerate}
\item
Let $I_1 = \{a,b\}$. 
After performing the fourth operation, we obtain
\begin{tabbing} 
\hspace{.2in} $P_2^{I_1}:\quad$ \= $\beta_A \vee \beta_B.$\\
\> $a \leftarrow \beta_A.$\\
\> $\beta_A \leftarrow a.$\\
\> $b \leftarrow \beta_B.$\\
\> $\beta_B \leftarrow b.$\\
\> $a \leftarrow b.$ 
\end{tabbing}
$P_2^{I_1}$ has only one minimal model, $M = \{a,\beta_A\}$; hence,
$I_1$ is not a stable model of $P_2$.
\item
Let $I_2 = \{a\}$. 
After performing the fourth operation, we obtain
\begin{tabbing} 
\hspace{.2in} $P_2^{I_2}:\quad$ \= $\beta_A.$\\
\> $a \leftarrow \beta_A.$\\
\> $\beta_A \leftarrow a.$\\
\> $a \leftarrow b.$ 
\end{tabbing}
$P_2^{I_2}$ has one minimal model, $M = \{a,\beta_A\}$; hence,
$I_2$ is a stable model of $P_2$.
\end{enumerate}
\end{example}

The introduction of disjunction into the head of a rule 
increases the expressiveness of the language, and allows
natural representation using disjunction. 

\begin{example} 
{\rm 
In scheduling, combinatorial counting or grouping is 
often needed. For example, a shift either has $a$ in it, or not.
If $a$ is in it, then either $a$ goes along with exactly one in $\{b,c\}$, or 
any two in $\{d,e,f\}$. 
This can be represented by a disjunctive program with 
cardinality constraints.

\begin{tabbing} 
\hspace{.2in} 
$1 \{a, not  \ a\} 1.$\\ 
\hspace{.2in} 
$1 \{b,c\} 1 \vee 2 \{d,e,f\} 2 \leftarrow a.$
\end{tabbing}
The semantics of this program can be understood
by the semantics of the corresponding constraint program:
\begin{tabbing}
\hspace{.2in} 
$(\{a\}, \{\emptyset,\{a\}\}).$\\
\hspace{.2in} 
$(\{b,c\}, \{\{b\},\{a\}\}) \vee (\{d,e,f\}, \{\{d,e\},\{d,f\},\{e,f\}\}) \leftarrow a.$
\end{tabbing}
This program has the following stable models: $\emptyset$, 
$\{a,b\}$, $\{a,c\}$, $\{a,d,e\}$, $\{a,d,f\}$, and
$\{a,f,e\}$. 
}
\end{example}

Once c-atoms are allowed to appear in the disjunctive head of a rule, 
disjunctive aggregates may be expressed.
\begin{example}
\label{sum}
{\rm 
Suppose the set of atoms in our propositional language is 
$\{p(-1), p(1), p(2)\}$.\footnote{Note that we assume a fixed propositional language 
that includes all the atoms appearing in
a given program.}
Consider the following program.

\begin{tabbing} 
\hspace{.2in} 
$p(1) \vee p(-1).$\\
\hspace{.2in} 
$\mbox{SUM}(X| p(X)) \ge 3 \vee \mbox{SUM}(X| p(X)) \le 0 \leftarrow \mbox{COUNT}(X|p(X)) \ge 1.$ 
\end{tabbing}
Its stable models are:  
$\{p(1),p(2)\}$,
$\{p(-1),p(1)\}$,
and
$\{p(-1)\}$.
}
\end{example}

As commented in \cite{SNS02},
a weight constraint can be transformed to one with negative weights 
but without negative literals. The 
weight constraints of this kind in fact express
linear inequations. Thus,
a disjunction of 
weight constraints can be viewed as 
specifying a disjunction of linear inequations.
For instance, the second rule in the above example can be 
expressed using weight constraints. To encode the SUM aggregate 
constraint above,
let
$l~ \Sigma~ u$ denote 
$l \{p(-1)=-1,p(1) = 1, p(2)=2\} u$, where $l$ and $u$ are the lower 
and upper bounds, respectively. When $l$ (resp. $u$) is omitted, 
it means $- \infty$ (resp. $\infty$). Then, we can write the following rule
$$3~ \Sigma \vee \Sigma~2 \leftarrow 1 \{p(-1)=1,p(1) = 1, p(2)=1\}$$
where the right hand side encodes the COUNT aggregate constraint.

We argue that 
disjunctive logic programming with constraint atoms provides a rich 
knowledge representation language for modeling
{\em conditional} as well as 
{\em disjunctive} constraints, which have been   
studied in the past 
in constraint programming
(see, e.g., \cite{disjunctive-constraint,disj-JACM,disjunctive-CSP}).\footnote{But note that disjunction in rule heads is 
{\em epistemic disjunction} \cite{GL91}, not the classic disjunction
in propositional logic.} 

\subsection{Properties of stable models}
We now show some properties of stable models.
\begin{theorem}
\label{th-smodel-model}
Any stable model $M$ of a logic program $P$ is a model of $P$.
\end{theorem}

A stable model may not be a minimal model for some constraint 
programs.
To illustrate, consider a logic program\\[.1in]
\hspace*{.2in} $P:\quad (\{a,b\}, \{\{a\} \uplus \{b\}, \{b\}\uplus \{a\}\}).$\\[.1in]
It is easy to check that $\{a\}$, $\{b\}$ and $\{a,b\}$ are all stable models of $P$.
We see that $\{a,b\}$ is not minimal.

It turns out that logic programs whose c-atoms appearing in rule heads
are elementary possess the minimality property.
 
\begin{theorem}
\label{th-smodel-1}
Let $P$ be a logic program such that c-atoms appearing in
the heads of its rules are all elementary. Any stable model of $P$
is a minimal model of $P$.
\end{theorem}

Recall that any atom $A$ can be expressed as a
c-atom $A' = (\{A\}, \{\{A\}\})$ and any 
negative literal $not\ A$ can be expressed as a c-atom
$A'' = (\{A\}, \{\emptyset\})$, such that for any interpretation $I$,
$I\models A$ (resp. $I\models not\ A$) if and only if 
$I\models A'$ (resp. $I\models A''$). The following
result further justifies our generalization of the standard
stable model semantics to logic programs with c-atoms.

\begin{theorem}
\label{th-smodel-2}
Let $P$ be a logic program with ordinary atoms
and $P'$ be $P$ with each positive literal $A$ replaced
by a c-atom $(\{A\}, \{\{A\}\})$, and each negative literal
$not\ A$ replaced by a c-atom $(\{A\}, \{\emptyset\})$.  
An interpretation $I$ is a stable model of $P$ if and only if 
it is a stable model of $P'$.
\end{theorem}

If all c-atoms are coded in the abstract representation,
the time complexity of the generalized Gelfond-Lifschitz transformation
is as follows.

\begin{theorem}
\label{th-complexity}
Let $P$ be a logic program with $n$ different 
c-atoms that are coded in the abstract representation
and $I$ be an interpretation.
Let $A$ be a c-atom
such that $I\models A$.
\begin{enumerate}
\item[{\rm (1)}]
The time complexity of computing all satisfiable sets of 
$A$ w.r.t. $T_A^I$ is linear in the size of $A_c^*$.

\item[{\rm (2)}]
The time complexity of the generalized Gelfond-Lifschitz transformation
is bounded by $O(|P| + n* (2M_{A_c^*} + M_{A_d} +1))$,
where $M_{A_c^*}$ and $M_{A_d}$ are the maximum sizes of $A_c^*$  and $A_d$ 
of a c-atom in $P$, respectively.
\end{enumerate}
\end{theorem}

The following result is immediate.

\begin{corollary}
\label{th-size}
The size of $P^I$ is bounded by
$O(|P| + n* (M_{A_c^*} + M_{A_d} + 1))$. 
\end{corollary}

Finally, we show the complexity of the major decision problem,
namely the stable model existence problem. 
In the following, we assume 
the explicit representation of c-atoms $A$ in the form
$(A_d,A_c)$ in a given program $P$.

\begin{theorem}
\label{th-complexity-2}
\begin{itemize}
\item [{\rm (1)}]
The problem of deciding  
whether 
a stable model exists for a normal constraint program $P$ is NP-complete. 
\item [{\rm (2)}]
The problem of deciding  
whether 
a stable model exists for a disjunctive constraint program $P$ is $\Sigma_P^2$-complete. 
\end{itemize}
\end{theorem}

\section{Relationship to Conditional Satisfaction}
\label{sec-Son}

Recently, \citeN{SPT06} proposed a fixpoint definition of 
stable models for logic programs with c-atoms. They introduce a key concept
termed {\em conditional satisfaction}.

\begin{definition}[\citeN{SPT06}]
\label{cond-sa}
Let $R$ and $S$ be two sets of atoms. The set $R$ conditionally
satisfies a c-atom $A$ w.r.t. $S$, denoted $R\models_S A$, 
if $R\models A$ and for every $S'$ such that $R\cap A_d\subseteq S'$
and $S'\subseteq S\cap A_d$, we have $S'\in A_c$.
\end{definition}

An immediate consequence operator $T_P(R, S)$ is introduced,
which evaluates each c-atom using the conditional
satisfaction $\models_S$ instead of the standard satisfaction
$\models$. 

\begin{definition}[\citeN{SPT06}]
\label{def-spt06}
Let $P$ be a positive basic logic program and 
$R$ and $S$ be two sets of atoms. Define
\[T_P(R,S) = \left\{A\left| \begin{array}{l}
                     \exists r\in P: R\models_S body(r),\\
                     head(r) = (\{A\}, \{\{A\}\})
                    \end{array}
                      \right. \right\}\]
\end{definition}

When the second argument is a model of $P$, $T_P$ is monotone
w.r.t. the first argument. In particular, given a model $M$ and let
$R\subseteq U\subseteq M$, then $T_P(R,M) \subseteq T_P(U,M) \subseteq M$.
Thus, for any model $I$, the sequence $T_P^i(\emptyset, I)$ with 
$T_P^0(\emptyset, I) = \emptyset$ and $T_P^{i+1}(\emptyset, I) = T_P(T_P^i(\emptyset, I), I)$,
converges to a fixpoint $T_P^\infty(\emptyset, I)$. $I$ is
defined to be a stable model if it is the same as the fixpoint. 

The following result reveals the relationship between
conditional satisfaction and satisfiable sets.

\begin{theorem}
\label{sat-set}
Let $A$ be a c-atom and $R$ and $I$ be two interpretations with $R\subseteq I$.
Let $T_A^I = I\cap A_d$.
$R\models_I A$ if and only if $A_c^*$ has an 
abstract prefixed power set $W\uplus V$ 
such that $R\cap A_d\uplus T_A^I \setminus  (R\cap A_d)$ 
is included in $W\uplus V$ (thus $W$ is a satisfiable set of
$A$ w.r.t. $T_A^I$ and $W\subseteq R\cap A_d$). 
\end{theorem}

Theorem \ref{sat-set} leads us to the conclusion that Son et al.'s fixpoint
definition and our definition of stable models are semantically equivalent
for positive basic programs,
as stated formally by the following theorem. 

\begin{theorem}
\label{equal-models}
Let $P$ be a positive basic program and $I$ a model of $P$.
$I$ is a stable model under Son et al.'s fixpoint definition
if and only if it is a stable model derived from the 
generalized Gelfond-Lifschitz transformation.
\end{theorem}

Note that by Theorem \ref{th-smodel-1}, any stable model of  
a positive basic program is a minimal model.

When the head $A$ of a rule $r$ is not elementary,
given an interpretation $I$, \citeN{SPT06} 
transform $r$ into the following set of rules:
\begin{tabbing}
$\qquad B\leftarrow body(r), \qquad\quad$ \= for each $B\in T_A^I$\\
$\qquad \bot \leftarrow B, body(r)$, \> for each $B\in F_A^I$.
\end{tabbing}
Under our generalized Gelfond-Lifschitz transformation,
$r$ is transformed into the following set of rules:
\begin{tabbing}
$\qquad \beta_A \leftarrow body(r)$, \\
$\qquad B \leftarrow \beta_A,\qquad\quad$ \=  for each $B\in T_A^I$\\
$\qquad \bot \leftarrow B, \beta_A$, \> for each $B\in F_A^I$\\
$\qquad \beta_A \leftarrow T_A^I$. 
\end{tabbing}
Apparently, the two transformations 
are semantically equivalent in that when $body(r)$ is true,
they derive the same conclusions except for the special atoms.
Combining with Theorem \ref{equal-models}, we then conclude 
that Son et al.'s fixpoint definition and our definition of stable models 
under the generalized Gelfond-Lifschitz 
transformation are semantically equivalent
for normal constraint programs.

Note that any normal constraint program can be transformed into a positive
basic program by replacing each negative literal $not\ B$ with
a c-atom $(\{B\}, \{\emptyset\})$ and 
replacing each negative c-atom $not\ A$ with
the complement $(A_d, 2^{A_d}\setminus  A_c)$ of $A$.
Therefore, our approach with the generalized
Gelfond-Lifschitz transformation is semantically equivalent to
Son et al.'s approach for normal constraint programs,
as stated by the following result.

\begin{corollary}
\label{equal-models-normal}
Let $P$ be a normal constraint program and $I$ a model of $P$.
Let $P'$ be $P$ with each negative literal $not\ B$ being replaced by
$(\{B\}, \{\emptyset\})$ and 
each negative c-atom $not\ A$ replaced by $(A_d, 2^{A_d}\setminus  A_c)$.
$I$ is a stable model of $P'$ under Son et al.'s approach
if and only if it is a stable model of $P'$ derived from the
generalized Gelfond-Lifschitz transformation.
\end{corollary}

\section{Properties based on Dependency Relation}
\label{properties}

In normal logic programming, the {\em dependency relation} 
over the atoms in 
a program is an essential notion based on which a number of important properties
are characterized (see, e.g., \cite{fages94,sato-dependency,y-y-94}). 
In this section, we extend these characterizations to 
normal constraint 
programs. A central question here is what 
should be the {\em dependency graph} for a given program. We will see that 
our abstract representation of c-atoms in the bodies of rules 
is precisely what is needed
to construct such a dependency graph, for the semantics defined 
by \citeN{SPT06}.

In this section, a {\em basic program} 
$P$ refers to 
a collection of rules of 
the
form
\begin{eqnarray} 
\label{basic-rule}
H \leftarrow A_1,...,A_n
\end{eqnarray}
where $H$ is either $\bot$ or an elementary c-atom, 
and $A_i$ are arbitrary c-atoms. Each rule in a basic program
is also called a {\em basic rule}.

To be consistent with the original definition of stable model \cite{GL88},
we assume that a rule of the form
$$\bot \leftarrow body$$
in a basic program 
is already 
replaced by 
a rule with an elementary head
$$f \leftarrow body, (\{f\},\{\emptyset\})$$
where $f$ is a new symbol representing the elementary c-atom 
$(\{f\},\{\{f\}\})$ and the c-atom 
$(\{f\},\{\emptyset\})$ in the body is 
its complement. 

The proof of the main result of this section is based on a method of 
representing 
a basic program by
a normal program, directly using the abstract representation of c-atoms,
while preserving the stable model semantics. 
Since the material is of 
interest on its own, we will first present it in the next subsection.

\subsection{Representing basic programs by normal programs}

The semantics of logic programs with c-atoms or aggregates 
have been studied by the unfolding approach \cite{PDB03,SP06-1}. It turns out,
under our abstract representation of c-atoms, the unfolding approach can be 
made simple.

Let $P$ be a basic program. The {\em normal program translation}
of $P$,
denoted $P_{n}$, is a normal program 
defined as follows.  
For each rule in $P$
\begin{eqnarray}
\label{unfolding-positive}
H \leftarrow A_1, ... , A_n
\end{eqnarray}
we have a rule 
\begin{eqnarray}
\label{extra-symbol}
H \leftarrow \theta_{A_1},...,\theta_{A_n}
\end{eqnarray}
in $P_{n}$, where $\theta_{A_i}$ are new symbols, plus
the following rules: for each $1 \leq i \leq n$, 
$$
\begin{array}{ll}
\theta_{A_i} \leftarrow W, not \ d_1, ..., not \ d_k~~~~~~\mbox{ for 
each $W \uplus V \in A_{i_c}^*$, 
where $\{d_1, ..., d_k\} = A_{i_d} \setminus W \cup V$.}
\end{array}
$$
\begin{example} 
{\rm Consider the program $P_2$ 
in Example \ref{prog-2} again, which consists of the following rules
$$
\begin{array}{ll}
p(1). ~~~~~~~~~~\\
p(-1) \leftarrow p(2).\\
p(2) \leftarrow \mbox{SUM}(\{X~|~p(X)\}) \geq 1.
\end{array}
$$
Let $A$ denote the aggregate in $P_2$. Recall that 
$A_c^* = \{\{p(1)\}\uplus\{p(2)\}, \{p(2)\}\uplus\{p(-1),p(1)\}\}$. 
Thus, $P_n$ consists of 
$$
\begin{array}{ll}
p(1).\\
p(-1) \leftarrow p(2). \\
p(2) \leftarrow \theta. \\ 
\theta \leftarrow p(1), not \ p(-1).\\
\theta\leftarrow p(2).
\end{array}
$$
It is clear that this normal program has no stable models.
}
\end{example} 

A distinct feature of our translation, as compared with 
the previous unfolding approach \cite{PDB03,SP06-1}, is that the abstract
representation of c-atoms is defined independently of any given program, while
in \cite{PDB03,SP06-1}, the translation to a normal program is an integrated 
process. This difference contributes to the simplicity of our approach.

The use of the abstract representation of c-atoms is essential. 
The following example shows that a simple enumeration of admissible 
solutions in a c-atom  
does not work. This is the case even for logic programs with only monotone
c-atoms.

\begin{example} {\rm 
Suppose a program $P$ that consists of a single rule
$$a \leftarrow A$$
where $A = (\{a\}, \{\emptyset, \{a\}\})$. Note that $A$ is monotone,
as well as a tautology, and 
$P$ has a unique stable model $\{a\}$. 
Since $A_c^* = \{\emptyset \uplus \{a\}\}$,
$P_n$ consists of 
$$
\begin{array}{ll}
a \leftarrow \theta.\\
\theta \leftarrow .
\end{array}
$$
Without the information encoded in the prefixed power set above, 
it may appear that a natural normal program encoding is to split 
admissible solutions as conditions into
different rules.
If we adopt this strategy, we will get the following normal program:
$$
\begin{array}{ll}
a \leftarrow \theta.\\
\theta \leftarrow not \ a.\\
\theta \leftarrow a.
\end{array}
$$
This program has no stable model.
}
\end{example} 

We now show that our translation preserves the stable models semantics.
Though the result is presented as a lemma for proving Theorem 
\ref{dependency-theorem} of the next subsection, 
it is obviously of independent interest.

Below, given a program $P$,
we denote by $ST(P)$ the set of stable models of $P$. 

\begin{lemma}
\label{positive-prog}
Let $P$ be a 
basic program and $P_n$ be its normal program translation.
Then,
$ST(P) = \{M_{|At(P)}| M \mbox{ is a stable model of }P_n\}$.
\end{lemma}

\subsection{Dependency relation-based characterizations}

We are now ready to extend some of the well-known
characterizations for normal programs to 
normal constraint programs. 
The key is the notion of a dependency graph for normal constraint programs. 

\begin{definition}
\label{dependency-graph}
Let $P$ be a basic program. The 
{\em dependency graph} $G_P$ is a graph $(V,E)$, where 
$V = At(P)$ and $E$ is the set of positive and negative
edges defined as the follows: 
there is a {\em positive edge} from $u$ to $v$, denoted 
$u \rightarrow^+ v$,
 if 
there is a rule $r$ of the form (\ref{basic-rule})
in $P$ such that $head(r) = u$, and for some $A_i \in body(r)$ and 
$W\uplus V \in A_{i_c}^*$,
$v \in W$;
there is a {\em negative edge} from $u$ to $v$,
denoted $u \rightarrow^- v$,
if 
there is a rule $r$ of the form (\ref{basic-rule})
in $P$ such that $head(r) = u$, and 
for some $A_i \in body(r)$ and 
$W\uplus V \in A_{i_c}^*$, 
$v \in A_{i_d} \setminus  W \cup V$.
\end{definition}

It is important to notice that, in the definition above,
for an abstract prefixed power set 
$W\uplus V \in A_{i_c}^*$, although 
we know that for any $I$ such that
$W \subseteq I \subseteq W \cup V$ we have  $I \in A_{i_c}$,  positive
edges are only into atoms in $W$, not into any atom in $I \setminus W$. 
Also, in normal logic programming,
negative edges are only into negative literals in rule bodies, but 
here a negative edge may result from a positive c-atom in the body of 
a rule.

\begin{example}
{\rm Suppose program $P$ consists of a single rule
$$
\begin{array}{ll} 
a \leftarrow (\{a,b,c\}, \{\emptyset, \{b\},\{b,c\} \}).
\end{array}
$$
Let $A$ be the c-atom in the body of the above rule.
Since $A_c^* = \{ \emptyset \uplus \{b\}, \{b\} \uplus \{c\} \}$,
we have $a \rightarrow^- a$, $a \rightarrow^- c$, and $a \rightarrow^+ b$. 
}
\end{example}

We say that $P$ has an {\em positive cycle} if 
there is a 
path in $G_P$ 
from an atom to itself via only positive edges. 
$P$ has an 
{\em odd cycle}
if 
there is a 
path in $G_P$ 
from an atom to itself via 
an odd number of negative edges, and $P$ has an {\em even cycle}
if 
there is a 
path in $G_P$ 
from an atom to itself via 
an even number of negative edges.
$P$ is said to be {\em call-consistent} if
$P$ has no odd cycles.
$P$ is {\em acyclic} if it has no cycle of any kind.   

We remark that our definition of dependency graph reduces to 
the standard one for normal programs.
Recall that the dependency graph for a normal program is defined as: 
for each normal rule in a normal program $P$
\begin{eqnarray}
\label{normal-rule}
a \leftarrow b_1,...,b_k, not \ c_1,..., not \ c_m
\end{eqnarray}
there is a positive edge $a \rightarrow^+ b_i$ in $G_P$ for each $i$, and 
a negative edge $a \rightarrow^- c_j$ for each $j$.

A normal program is in fact a basic program, in the sense that
each positive literal $b_i$ in the rule above
is replaced by an elementary
c-atom $(\{b_i\}, \{\{b_i\}\})$ and 
each negative literal $not \ c_i$ replaced by 
$(\{c_i\}, \{\emptyset\})$, i.e., the complement of 
$(\{c_i\}, \{\{c_i\}\})$.  
Let
the resulting program be $P'$.
Since if 
$C_i= (\{c_i\}, \{\emptyset\})$ then 
$C_{i}^*= (\{c_i\}, \{\emptyset \uplus \emptyset\})$,
by Definition \ref{dependency-graph},
there is 
a negative edge $a \rightarrow^- c_i$ in $G_{P'}$. 

The following theorem shows that 
the well-known properties based on the dependency graphs 
for normal programs as shown in \cite{y-y-94}
remain to hold for normal constraint programs 
under the new definition of dependency graph for the latter.
\begin{theorem}
\label{dependency-theorem}
Let $P$ be a basic program.
\begin{enumerate}
\item [{\rm (1)}] 
$P$ has a stable model if $P$ is call-consistent. 
\item [{\rm (2)}]
$P$ has more than one stable model only if $P$ has an even loop.
\item [{\rm (3)}]
$P$ has a unique stable model if $P$ is acyclic.
\item [{\rm (4)}]
If $P$ has no positive cycles, then every supported model of 
$P$ is a stable model of $P$.
\end{enumerate}
\end{theorem}

\begin{example}
{\rm To illustrate the point (2) above,
consider the following program.
$$
\begin{array}{ll}
p \leftarrow.\\
a \leftarrow (\{p,b\}, \{\{p\}\}).\\
b \leftarrow (\{p,a\}, \{\{p\}\}).\\
\end{array}
$$
The program has two stable models $\{p,a\}$ and $\{p,b\}$. Then, according to
the theorem, there must exist an even loop in its dependency graph. Indeed, 
the edges $a \rightarrow^- b$ and 
$b \rightarrow^- a$ form 
such an even cycle.
}
\end{example}

\section{Related Work}
\label{sec-rel}
The notion of logic programs with c-atoms is introduced in \cite{MR04,MT04}
and further developed in \cite{LT05,MNT06,SP06-1,SPT06}.
As we mentioned earlier, major existing approaches can be roughly classified into
three types: unfolding approaches,
fixpoint approaches, and minimal model approaches. 

Representative {\em unfolding} approaches to handling c-atoms include 
\cite{PDB03,SP06-1}, where a notion of aggregate solutions (or solutions)
is introduced. Informally, a solution of a c-atom 
$A = (A_d, A_c)$ is a pair $\langle S_1,S_2\rangle$
of disjoint sets of atoms of $A_d$ such that for every interpretation $I$,
if $S_1\subseteq I$ and $S_2\cap I = \emptyset$ then 
$I\models A$. This definition
is given by \citeN{SP06-1}.
\citeN{PDB03} define an aggregate solution $A$
as a pair $\langle S_1,S_2\rangle$ with $S_1\subseteq S_2\subseteq A_d$
such that for every interpretation $I$,
if $S_1\subseteq I$ and $(A_d\setminus S_2)\cap I = \emptyset$ then $I\models A$.
In the following, we use the former definition.

It turns out that each $W\uplus V\in A_c^*$ corresponds to
a minimal solution $\langle W, A_d \setminus  (W\cup V) \rangle$ of $A$.
A solution $\langle S_1,S_2\rangle$ of $A$ is minimal if 
for no $S_3\subset S_1$ nor $S_4\subset S_2$, 
$\langle S_3,S_2\rangle$ or $\langle S_1,S_4\rangle$ 
is a solution of $A$. Firstly, $\langle W, A_d \setminus  (W\cup V) \rangle$ 
is a solution of $A$; by Theorem \ref{sem-catom-2}
for any interpretation $I$,
if $W\subseteq I$ and $(A_d \setminus  (W\cup V)) \cap I = \emptyset$ then 
$I\models A$. Secondly, $\langle W, A_d \setminus  (W\cup V) \rangle$ 
is a minimal solution of $A$, as by Theorem \ref{sem-catom-3},
$W\wedge not\ (A_d \setminus  (W\cup V))$ cannot be further
simplified.

Representative {\em fixpoint} approaches include \cite{LPST07,MNT06,MT04,Pel04,PT04,SPT06,SPT07}.
Son et al. \citeyear{SPT06,SPT07} can handle arbitrary c-atoms, while \cite{MNT06,MT04,PT04}
apply only to monotone c-atoms. \citeN{LPST07} extend \cite{LT05,MNT06,MT04,PT04}
for arbitrary c-atoms based on a concept of {\em computation}.
Son et al. \citeyear{SPT06,SPT07} show that 
their fixpoint approach is semantically equivalent to 
that of \citeN{MT04} for normal logic programs
with monotone c-atoms; equivalent to that of \citeN{FLP04} 
and \citeN{Ferr05} for positive basic logic programs 
with monotone c-atoms; equivalent to that of \cite{DPB01,PDB03} 
for positive basic logic programs with arbitrary c-atoms. 
In Section \ref{sec-Son}, we show that our approach using
the generalized Gelfond-Lifschitz transformation
is semantically equivalent to Son et al.'s approach for normal 
logic programs with arbitrary c-atoms.
Therefore, the stable model semantics defined in this paper
for disjunctive logic programs with arbitrary c-atoms 
extends these existing semantics.

\citeN{FLP04} propose a {\em minimal model} approach.
To check if an interpretation $I$ is a stable model of $P$, 
they first remove all rules in $P$ whose bodies are not satisfied by $I$, 
then define $I$ to be a stable model 
if it is a minimal model of the simplified program. 
They consider the class of disjunctive logic programs 
whose rule heads are a disjunction of ordinary atoms.
Stable models of $P$ under this semantics are minimal models of $P$.
\citeN{Ferr05} defines a stable model semantics 
in a different way, which 
(when negated c-atoms are treated as their complement c-atoms)
agrees with the 
minimal-model based one on this class of programs. 
\citeN{SPT06} show that for normal logic programs 
whose c-atoms appearing in rule heads are elementary,
stable models under their semantics are stable models under the semantics
of \citeN{FLP04} and \citeN{Ferr05}.  
It immediately follows that for such normal logic programs,
stable models under our semantics are stable models under the semantics
of Faber et al. and Ferraris.  
However, the converse is not necessarily true,
even for positive basic logic programs. 
Consider the positive basic logic program $P$:
\begin{tabbing} 
\hspace{.2in} \= $b\leftarrow c.$\\
\>               $c \leftarrow d.$\\
\>               $d \leftarrow  (\{b,c\}, \{\emptyset, \{b\}, \{b,c\}\}) .$
\end{tabbing}
$P$ has only one model $I = \{b,c,d\}$.
It is easy to check that $I$ is not a stable model under 
the semantics of \citeN{SPT06} and ours. 
However, $I$ is a stable model under the semantics
of \citeN{FLP04} and \citeN{Ferr05}.  
Observe that the truth of $b,c, d$
can only be inferred via a self-supporting loop: 
\[b\rightarrow d \rightarrow c \rightarrow b.\]
This example program indicates that both the semantics
of Faber et al. and that of Ferraris 
allow self-supporting loops. 

\section{Conclusions and Future Work}
\label{sec-concl}

In this paper 
we have introduced  
an abstract representation 
of c-atoms. 
To substantiate the claim that the abstract representation captures the 
essential information correctly and compactly, we showed two applications.
In the first one,
we show that the semantics based on conditional satisfaction 
\cite{SPT06,SPT07}, and the one equivalent to it \cite{DPB01}, can
be defined by a generalized form of Gelfond-Lifschitz transformation,
thus demonstrating that 
Gelfond-Lifschitz transformation can still play an important role 
in the study of semantics for logic programs with arbitrary c-atoms.
In the second application,
we show that our abstract representation
of c-atoms encodes the information needed to define the atom dependency 
relation in a given program. As a result, the properties known to normal programs
can be extended to programs with c-atoms. In this process,
the unfolding approach \cite{SP06-1}
is made simple.

Several interesting tasks remain open. One is the possibility of 
showing that 
other semantics may be characterized by our abstract
representation of c-atoms. This is because 
prefixed power sets identify ``monotone components'' 
of c-atoms.
Another task is to develop new algorithms for efficiently
constructing the abstract form of c-atoms from the power set form representation. 
Finally, methods for computing the stable models 
(under our generalized Gelfond-Lifschitz transformation) of 
logic programs with arbitrary c-atoms remain a challenging open problem. 

\section{Acknowledgments}
We would like to thank the anonymous referees for their constructive comments and 
suggestions that helped us improve this work. 
Yi-Dong Shen is supported in part by 
NSFC grants 60673103,
60721061 and 60833001, and by the National High-tech R\&D Program (863 Program).
The work by Jia-Huai You and Li-Yan Yuan is supported in part by 
the Natural Sciences and Engineering Research Council of Canada.

\appendix
\section{Proof of Theorems and Lemmas}

\noindent {\bf Proof of Theorem \ref{t1}:} 
Assume that $I\uplus J$ is included in $I_1\uplus J_1$.
We first prove $I_1\subseteq I$. 
If on the contrary $I_1\not\subseteq I$, 
there is an atom $a$ such that $a\in I_1$ and $a\not\in I$.
This means that every $S$ covered by $I_1\uplus J_1$
must contain $a$. Since $I$ is covered by $I\uplus J$, $I$ is covered by $I_1\uplus J_1$.
But $I$ does not contain $a$, a contradiction. 
We now prove $I\cup J\subseteq I_1\cup J_1$. 
If on the contrary $I\cup J\not\subseteq I_1\cup J_1$, 
$I\cup J$ is not covered by $I_1\uplus J_1$. This means    
$I\uplus J$ is not included in $I_1\uplus J_1$, a contradiction.

Next, assume that $I\uplus J$ is included in $I_1\uplus J_1$
and $I_1\uplus J_1$ is included in $I_2\uplus J_2$.
We have $I_2\subseteq I_1\subseteq I$, and  
$I\cup J\subseteq I_1\cup J_1 \subseteq I_2\cup J_2$. 
This means all sets covered by $I\uplus J$ 
are covered by $I_2\uplus J_2$.
That is, $I\uplus J$ is included in $I_2\uplus J_2$.  \hfill \qed\\[.1in]
\noindent {\bf Proof of Theorem \ref{ab-form}:} 
(1) For each $S\in A_c$, the collection $C_S$ of 
abstract $S$-prefixed power sets of $A$ is uniquely defined by Definition \ref{ab-1},
thus $A_c^*$ is uniquely defined by Definition \ref{ab-2}.

(2) Assume $I\models A$, i.e., $I\cap A_d = S\in A_c$.
By Definition \ref{ab-1}, 
the collection $C_S$ of abstract $S$-prefixed power sets of $A$
contains $S\uplus S_i$ covering $S$.
By Definition \ref{ab-2}, $A_c^*$ has an 
abstract prefixed power set $W\uplus V$ such that
either $W\uplus V = S\uplus S_i$ or 
$S\uplus S_i$ is included in $W\uplus V$. 
This means that $W\uplus V$ covers $S$.

Conversely, assume that $A_c^*$ has an 
abstract prefixed power set $W\uplus V$ covering $I\cap A_d$. 
By Definition \ref{ab-2}, $W\uplus V$ is an 
abstract $W$-prefixed power set of $A$ with $W\in A_c$.
By Definition \ref{ab-1}, all sets covered by $W\uplus V$ are in $A_c$.
This means $I\cap A_d\in A_c$, and hence $I\models A$.   \hfill \qed\\[.1in]
\noindent {\bf Proof of Theorem \ref{mono-conv-catom}:} 
Let $G = \bigcup_{S\in A_c} C_S$, where $C_S$ is the
collection of abstract $S$-prefixed power sets of $A$.
By Definition \ref{ab-2}, $A_c^*$
is $G$ with all redundants removed.
\begin{enumerate}
\item[{\rm (1)}] 
($\Longrightarrow$)
Assume that $A$ is monotone. 
Then, all supersets of $S\in A_c$ from $2^{A_d}$ are in $A_c$,
so all abstract $S$-prefixed power sets in $G$
must be of the form $S\uplus A_d\setminus S$.
If $S$ is not minimal in $A_c$, $S\uplus A_d\setminus S$
is redundant in $G$ since for some $S'\subset S$, which is minimal in $A_c$, 
$S'\uplus A_d\setminus S'$ is in $G$.
Therefore, $A_c^* =\{B \uplus A_d \setminus B :  \ B \mbox{ is minimal in } A_c\}$.
Clearly, $|W| + |V| = |A_d|$
for each $W\uplus V\in A_c^*$. 

($\Longleftarrow$) Assume that for every $W\uplus V\in A_c^*$,
we have $|W| + |V| = |A_d|$; i.e. $V = A_d\setminus W$. 
Every abstract $S$-prefixed power set in $G$
must be of the form $S\uplus A_d\setminus S$,
for otherwise, there is one $W\uplus V\in A_c^*$ with
$W\subseteq S$ and $V \subset A_d\setminus W$.
As shown above, in this case every 
$S\uplus A_d\setminus S$ in $G$ 
is redundant unless $S$ is minimal in $A_c$. 
Therefore, $A_c^*$ is $G$ with all $S\uplus A_d\setminus S$
removed, where $S$ is not minimal in $A_c$. 
That is, $A_c^* =\{B \uplus A_d \setminus B :  \ B \mbox{ is minimal in } A_c\}$.
This shows that for any $S'$ which is minimal in $A_c$,
all supersets of $S'$ are in $A_c$.
For any $S\in A_c$, there is some $S'\subseteq S$, which is minimal in $A_c$.
Since all supersets of $S'$ are in $A_c$, 
all supersets of $S$ are in $A_c$.
This shows that $A$ is monotone.

\item[{\rm (2)}] 
($\Longrightarrow$)
Assume that $A$ is antimonotone. 
Every abstract $S$-prefixed power set in $G$
must be of the form $\emptyset \uplus T$.
By Definition \ref{ab-1}, $T$ is maximal in $A_c$.
That is,
$A_c^* =\{\emptyset \uplus T  :  \ T \mbox{ is maximal in } A_c\}$.
Clearly,
$W = \emptyset$ 
for each $W\uplus V\in A_c^*$. 

($\Longleftarrow$) Assume that every abstract prefixed power set in $A_c^*$
is of the form $\emptyset \uplus T$. 
By Definition \ref{ab-1}, $T$ is maximal in $A_c$.
That is,
$A_c^* =\{\emptyset \uplus T  :  \ T \mbox{ is maximal in } A_c\}$.
Clearly,
for any $T\in A_c$ all subsets of $T$ are in $A_c$.
This shows that $A$ is antimonotone. 
 
\item[{\rm (3)}] 
($\Longrightarrow$)
Assume that $A$ is convex.
Consider $B\uplus T$ in $G$.
If $B$ is not minimal in $A_c$,
since $A$ is convex $B\uplus T$ is included in
$B'\uplus T$, where $B'\subset B$ is minimal in $A_c$.
For the same reason, 
if $B\cup T$ is not maximal in $A_c$,
$B\uplus T$ is included in
$B\uplus T'$, where $T'\supset T$ and $B\cup T'$ is maximal in $A_c$.
In both cases, $B\uplus T$ 
is redundant in $G$.
Therefore, $A_c^*$ is $G$ with all $B\uplus T$ 
removed, where either $B$ is not minimal or $B\cup T$ is not maximal in $A_c$. 
That is, 
$A_c^* =\{B \uplus T   : \ 
B  \mbox{ is minimal and } B\cup T \mbox{ is maximal in } A_c\}$.  

($\Longleftarrow$) Assume
$A_c^* =\{B \uplus T   : \ 
B  \mbox{ is minimal and } B\cup T \mbox{ is maximal in } A_c\}$. 
Then, for any $S_1,S_2\in A_c$ with $S_1\subset S_2$,
there is some $B \uplus T$ in $A_c^*$,
which covers all $S$ with $B\subseteq S_1\subseteq S \subseteq S_2 \subseteq B\cup T$.
This means that all subsets in between $S_1$ and $S_2$ 
are in $A_c$. That is,
$A$ is convex.
\end{enumerate}
\hfill \qed\\[.1in]
\noindent {\bf Proof of Theorem \ref{construction-abstract-rep}:}
Let $A$ be a c-atom. 
We use a simple algorithm to construct $A^*_c$.
The algorithm returns a set, say $\Pi$,
which is set to $\emptyset$ at the beginning.

Note that for any $p\in A_c$, $|p|\leq |A_d|$.
Therefore, for any $p,q \in A_c$ it takes 
$O(|A_d|^2)$ time to determine if $p$ is a subset of $q$.
Moreover, when $p \subset q$, there are at most
$2^{|q\setminus p|}-2$ sets $w$ such that
$p \subset w \subset q$.

For each pair $(p,q)$, 
where $p,q \in A_c$ and $p \subset q$,
let $S$ be the set of all $w \in A_c$ such that 
$p \subset w \subset q$.
If $|S| = 2^{|q\setminus p|}-2$, we add 
$p \uplus q \setminus p$ to $\Pi$.
Since there are at most $O(|A_c|^2)$ such pairs to check, 
and for each, it takes 
$O(|A_c| * |A_d|^2)$ time to perform the test
(i.e., for each $w\in A_c$ we check if $p \subset w \subset q$),
the time for the above process is bounded by $O(|A_c|^3 * |A_d|^2)$. Note that
$|\Pi|$ is bounded by $O(|A_c|^2)$.

After the above process, all possible 
abstract prefixed power sets of $A$
are in the resulting $\Pi$. 
Then, we remove all (redundant) $\pi$ from $\Pi$ if 
$\pi$ is included in some other $\xi \in \Pi$. 
By Theorem \ref{t1}, it takes $O(|A_d|^2)$ time
to check if 
$\pi$ is included in $\xi$.  
Therefore, the time for this redundancy removing process is bounded by
$O(|A_c|^4 * |A_d|^2)$.

As a result, $\Pi$ consists of all non-redundant 
abstract prefixed power sets of $A$.
By Definition \ref{ab-2}, $\Pi$ is $A_c^*$.
In total, it takes $O(|A_c|^4 * |A_d|^2)$ time 
to construct $A^*$ from $A$. \hfill \qed\\[.1in]
\noindent {\bf Proof of Proposition \ref{th-sem}:} 
(1) Assume that $I$ satisfies $A$; i.e., $A_d\cap I = S_i\in A_c$.
Then, we have $C_i = S_i \wedge not\ (A_d \setminus  S_i)$ with
$S_i\subseteq I$ and $(A_d \setminus  S_i) \cap I = \emptyset$.
This means that both $S_i$ and $not\ (A_d \setminus  S_i)$ are
true in $I$. Hence, $C_i$ is true in $I$ and thus   
$C_1\vee ...\vee C_m$ is true in $I$.
Conversely, assume that $C_1\vee ...\vee C_m$ is true in $I$.
Some $C_i = S_i \wedge not\ (A_d \setminus  S_i)$ must be true in $I$,
meaning that $S_i\subseteq I$ and $(A_d \setminus  S_i) \cap I = \emptyset$.
This shows that $A_d\cap I = S_i$. Since $S_i\in A_c$, $I$ satisfies $A$.

(2) Assume that $I$ satisfies $not\ A$; i.e., $A_d\cap I\not\in A_c$.
Then, every $C_i = S_i \wedge not\ (A_d \setminus  S_i)$
is false in $I$ because either $S_i\not\subseteq I$ 
or $(A_d \setminus  S_i) \cap I \neq \emptyset$.
Thus, $not\ (C_1\vee ...\vee C_m)$ is true in $I$.
Conversely, assume that $not\ (C_1\vee ...\vee C_m)$ is true in $I$;
i.e., every $C_i = S_i \wedge not\ (A_d \setminus  S_i)$ is false in $I$.
This means that for each $S_i\in A_c$, either $S_i\not\subseteq I$ 
or $(A_d \setminus  S_i) \cap I \neq \emptyset$; therefore,
$A_d\cap I \neq S_i$. This shows
$A_d\cap I\not\in A_c$; thus
$I$ satisfies $not\ A$. \hfill \qed\\[.1in]
\noindent {\bf Proof of Lemma \ref{catom-lem}:} 
The proof is by induction on $k$ with $1\leq k\leq m$.
When $k=1$ (induction basis), $F=a_1 \vee not\ a_1 \equiv true$.
For the induction hypothesis, assume that 
$F = \bigvee_{1\leq i\leq k,\  L_i\in \{a_i, not\ a_i\}} L_1\wedge ...\wedge L_k$
can be simplified to $true$
by applying rule (\ref{s-rule})
for any $k<m$. This holds for $k=m$, as shown below:
\begin{tabbing}
$\qquad F\ $\= $=$ \= $\bigvee_{1\leq i\leq m,\  L_i\in \{a_i, not\ a_i\}} L_1\wedge ...\wedge L_m$ \\ 
\> $\equiv [\bigvee_{1\leq i\leq (m-1),\  L_i\in \{a_i, not\ a_i\}} (L_1\wedge ...\wedge L_{m-1})\wedge a_m]\ \vee$ \\
\>\> $[\bigvee_{1\leq i\leq (m-1),\  L_i\in \{a_i, not\ a_i\}} (L_1\wedge ...\wedge L_{m-1})\wedge not\ a_m]$\\   
\> $\equiv a_m\wedge [\bigvee_{1\leq i\leq (m-1),\  L_i\in \{a_i, not\ a_i\}} L_1\wedge ...\wedge L_{m-1}]\ \vee$ \\
\>\> $not\ a_m \wedge [\bigvee_{1\leq i\leq (m-1),\  L_i\in \{a_i, not\ a_i\}} L_1\wedge ...\wedge L_{m-1}]$\\   
\> $\equiv a_m \vee not\ a_m$ (by the induction hypothesis)\\
\> $\equiv true$ \`\qed
\end{tabbing}
\noindent {\bf Proof of Theorem \ref{sem-catom-2}:} 
By Theorem \ref{ab-form}, $A_c$ and $A_c^*$ express the same set
of admissible solutions to $A$ in that for any $S\subseteq A_d$, 
$S\in A_c$ if and only if $A_c^*$ contains an abstract prefixed 
power set $W\uplus V$ covering $S$. 
Let $V=\{a_1,...,a_m\}$.
Note that each $W\uplus V$ in $A_c^*$ 
exactly covers the set $\{W\cup S|S\subseteq V\}$ 
of items in $A_c$, and all items in $A_c^*$ 
exactly cover all items in $A_c$. 
Since the semantics of each $S\in A_c$ 
is $S \wedge not\ (A_d \setminus  S)$,
the semantics of each $W\uplus V$ in $A_c^*$ is
\begin{tabbing}
$\qquad\qquad$ \= $\bigvee_{1\leq i\leq m,\ L_i\in \{a_i, not\ a_i\}} W\wedge (L_1\wedge ...\wedge L_m)\wedge not\ (A_d \setminus  (W\cup V))$ \\ 
\> $ \equiv W\wedge not\ (A_d \setminus  (W\cup V)) \wedge [\bigvee_{1\leq i\leq m,\ L_i\in \{a_i, not\ a_i\}} L_1\wedge ...\wedge L_m]$ 
\end{tabbing}
which, by Lemma \ref{catom-lem}, can be simplified 
to $W\wedge not\ (A_d \setminus  (W\cup V))$
by applying rule (\ref{s-rule}).
Thus, we have
\begin{tabbing}
$\qquad\qquad A$ \= $\equiv \bigvee_{S \in A_c} S \wedge not\ (A_d \setminus  S)$\\
\>                   $\equiv \bigvee_{W\uplus V \in A_c^*} \bigvee_{1\leq i\leq m,\ \ L_i\in \{a_i, not\ a_i\}} W\wedge (L_1\wedge ...\wedge L_m)\wedge not\ (A_d \setminus  (W\cup V))$\\
\>                   $\equiv \bigvee_{W\uplus V \in A_c^*} W \wedge not\ (A_d \setminus  (W\cup V))$
\end{tabbing}
\hfill \qed\\[.1in]
\noindent {\bf Proof of Theorem \ref{sem-catom-3}:} 
For any two $W_1 \uplus V_1, W_2 \uplus V_2\in A_c^*$,
we distinguish between three cases:
(1) if $W_1 = W_2$, then the two conjunctions
$W_1 \wedge not\ (A_d \setminus  (W_1\cup V_1))$ and $W_2 \wedge not\ (A_d \setminus  (W_2\cup V_2))$
have no conflicting literals, thus they cannot be pairwise
simplified using rule (\ref{s-rule});
(2) if $W_1 \subset W_2$ with $|W_2| - |W_1| = 1$, then $V_1\neq V_2$ 
(otherwise, $W_1 \uplus V_1\cup (W_2\setminus  W_1)$ 
should be in $A_c^*$ so that $W_1 \uplus V_1$ is not in $A_c^*$),
which means that the two conjunctions
$W_1 \wedge not\ (A_d \setminus  (W_1\cup V_1))$ and $W_2 \wedge not\ (A_d \setminus  (W_2\cup V_2))$
have at least two different literals, one in their positive part and another in their negative part,
so that they cannot be pairwise
simplified using rule (\ref{s-rule});
(3) otherwise (i.e., $W_1\neq W_2$ and
$W_1 \not\subset W_2$ and $W_2 \not\subset W_1$, or 
$W_1 \subset W_2$ with $|W_2| - |W_1| > 1$), 
the two conjunctions
$W_1 \wedge not\ (A_d \setminus  (W_1\cup V_1))$ and $W_2 \wedge not\ (A_d \setminus  (W_2\cup V_2))$
have at least two different positive literals, thus they cannot be pairwise
simplified using rule (\ref{s-rule}).
\hfill \qed\\[.1in]
\noindent {\bf Proof of Theorem \ref{catom-chara}:} 
By Proposition \ref{th-sem}, $I\models A$ if and only if $I \models
\bigvee_{S\in A_c} S \wedge not\ (A_d \setminus  S)$, and 
by Theorem \ref{sem-catom-2}, 
if and only if $I$ satisfies 
$\bigvee_{W\uplus V\in A_c^*} W \wedge not\ (A_d \setminus  (W\cup V))$.
For each $W\uplus V\in A_c^*\setminus  A_s^I$, since it does not cover $T_A^I$,
$W \wedge not\ (A_d \setminus  (W\cup V))$ is false in $I$.
This means that $\bigvee_{W\uplus V\in A_c^*} W \wedge not\ (A_d \setminus  (W\cup V))$
is true in $I$ if and only if 
$\bigvee_{W\uplus V\in A_s^I} W \wedge not\ (A_d \setminus  (W\cup V))$ is true in $I$.
Therefore, $I\models A$ if and only if $I \models
\bigvee_{W\uplus V\in A_s^I} W \wedge not\ (A_d \setminus  (W\cup V))$.
\hfill \qed \\[.1in]
\noindent {\bf Proof of Theorem \ref{th-num-min}:} 
When $S$ is a satisfiable set, there is an 
abstract $S$-prefixed power set $S\uplus S_1$ in $A_c^*$
such that $T_A^I$ is covered by $S\uplus S_1$. By the definition
of an abstract prefixed power set, every $S'$ 
with $S\subseteq S'\subseteq T_A^I$ is covered by $S\uplus S_1$.  
By Definition \ref{ab-1}, every such $S'$ is in $A_c$. \hfill \qed\\[.1in]
\noindent {\bf Proof of Theorem \ref{th-smodel-model}:} 
Let $M$ be a stable model of $P$
obtained by applying the generalized Gelfond-Lifschitz transformation.
Note that $\bot$ is not in $M$.
To prove that $M$ is a model of $P$ is to prove that for any rule $r$ in $P$ 
we have $M\models r$. By definition, if $M\models head(r)$ or $M\not\models body(r)$ then $M\models r$. 
Assume that $M\not\models head(r)$ and, on the contrary, that $M\models body(r)$.
Let $r$ take the form
\[H_1\vee ... \vee H_k \leftarrow B_1,...,B_m,A_1,...,A_n, not\ C_1, ..., not\ C_l\]
where each $B_i$ or $C_i$ is an atom and each $A_i$ is a c-atom. 
$H_i$ can be an atom or a c-atom. We then have $M\models B_i$,
$M\models A_i$, $M\models not\ C_i$ and $M\not\models H_i$. 

For every negative literal $not\ C_i$ in $body(r)$, since $M\models not\ C_i$ it 
will be removed in the second operation of the generalized Gelfond-Lifschitz 
transformation. For every c-atom $A_i$ in $body(r)$, since $M\models A_i$ 
it will be replaced in the third operation by a special atom $\theta_{A_i}$ along with
a new rule $\theta_{A_i}\leftarrow D_1, ..., D_t$ for each 
satisfiable set $\{D_1, ..., D_t\}$ of $A_i$ w.r.t. $T_{A_i}^M$.
As a result, the generalized Gelfond-Lifschitz 
transformation $P^M$ contains the following rules derived from $r$:
\begin{tabbing} 
$H_1'\vee ... \vee H_k'\leftarrow B_1,...,B_m,\theta_{A_1},...,\theta_{A_n},$  \\ 
$\theta_{A_i} \leftarrow D_1, ...,$ \= $D_t,$ $\ $ for each 
c-atom $A_i$ and each\\
\>  satisfiable set $\{D_1, ..., D_t\}$ of $A_i$ w.r.t. $T_{A_i}^M$  
\end{tabbing}
Here, $H_i'$ is $H_i$ if $H_i$ is an atom; or when $H_i$ is a c-atom, $H_i'$ is $\bot$ 
because $M\not\models H_i$ ($H_i$ is replaced by $\bot$ in the fourth operation).

Let $N$ be a minimal model of $P^M$ with $M=N\setminus  \Gamma$
(which leads to $M$ being a stable model of $P$).   
For each c-atom $A_i$, we have $M\cap A_{i_d} = N\cap A_{i_d} = T_{A_i}^M$.
Since each satisfiable set $\{D_1, ..., D_t\}$ of $A_i$ 
is a subset of $T_{A_i}^M$, we have $\{D_1, ..., D_t\}\subseteq M \subseteq N$.
This means that for each $A_i$, the body of the rule
\[\theta_{A_i} \leftarrow D_1, ..., D_t\]     
in $P^M$ is satisfied in $N$. Since $N$ is a minimal model of $P^M$,
the head $\theta_{A_i}$ of the above rule must be in $N$.
As a result, the body of the rule  
\[H_1'\vee ... \vee H_k'\leftarrow B_1,...,B_m,\theta_{A_1},...,\theta_{A_n}\]  
in $P^M$ is satisfied in $N$, thus some $H_j'$ in the head is in $N$.
Since no $H_i'$ is a special atom prefixed with $\theta$ or $\beta$,
$H_j'$ is also in $M$. Since $\bot$ is not in $M$, $H_j'$ must be $H_j$ in 
the rule $r$. This means that $M$ satisfies $head(r)$,  
contradicting the assumption $M\not\models head(r)$.
We then conclude that $M$ is a model of $P$.
\hfill \qed \\[.1in]
\noindent {\bf Proof of Theorem \ref{th-smodel-1}:} 
Let $I$ be a stable model of $P$ and $M$ be a minimal model 
of the generalized Gelfond-Lifschitz transformation
$P^I$ with $I = M\setminus \Gamma$. 
Let $(P^I)^i$ be obtained from $P$ after performing the $i$-th
operation ($i = 1,...,4$) in Definition \ref{gen-smodel}.
Note that $P^I = (P^I)^4$. 

Since every c-atom $A$ appearing in each rule head is an elementary c-atom
of the form $(\{a\}, \{\{a\}\})$, the semantics of $P$ will not be changed if
we replace $A$ in the head with a new symbol $\beta_A$
and define $\beta_A$ by the two rules $\beta_A\leftarrow a$
and $a\leftarrow \beta_A$ (expressing $\beta_A\equiv a$). 
This means that when c-atoms in the rule heads are all elementary,
performing the fourth operation in Definition \ref{gen-smodel}
does not change the semantics of $P$. Therefore,
since $M$ is a minimal model of $(P^I)^4$,
$M\setminus \Gamma_\beta$ is a minimal model of $(P^I)^3$. 

Note that for each rule (introduced in the third operation) of the form  $\theta_A\leftarrow W$,
where $W=\{A_1,...,A_m\}$ is a satisfiable set, we have $W\subset M\setminus \Gamma_\beta$ 
and $\theta_A\in M\setminus \Gamma_\beta$. Let $Q$ be the set of rules in $(P^I)^3$
whose heads are not special atoms prefixed with $\theta$. For any non-empty set $S$ of $M\setminus \Gamma$,
$M\setminus (\Gamma_\beta \cup S)$  will not satisfy $Q$; otherwise, 
$M\setminus \Gamma_\beta$ 
would not be a minimal model of $(P^I)^3$. 

Let $Q_1$ be $(P^I)^3$ such that all rules $\theta_A\leftarrow W$ with 
the same head $\theta_A$ are replaced by a compact rule 
\[\theta_A\leftarrow \bigvee_{W\uplus V \in A_c^*} W \wedge not\ (A_d \setminus (W\cup V))\]
Since $M\setminus \Gamma_\beta$ is a minimal model of $(P^I)^3$,
$M\setminus \Gamma_\beta$ is a minimal model of $Q_1$.

Now let $Q_2$ be $Q_1$ obtained by first replacing 
all occurrences of each $\theta_A$ in rule bodies 
with the body of the above compact rule,
then removing all compact rules. 
Since $M\setminus \Gamma_\beta$ is a minimal model of $Q_1$,
$M\setminus \Gamma$ is a minimal model of $Q_2$.
Note $I = M\setminus \Gamma$.

By Theorem \ref{sem-catom-2}, we can replace each 
$\bigvee_{W\uplus V \in A_c^*} W \wedge not\ (A_d \setminus  (W\cup V))$
in $Q_2$ with c-atom $A$ without changing the semantics of $Q_2$.
This transforms $Q_2$ into $(P^I)^2$.  
Therefore, $I$ is a minimal model of $(P^I)^2$. 

$(P^I)^2$ is $(P^I)^1$ with all negative literals removed.  
Since all such negative literals are satisfied by $I$,
that $I$ is a minimal model of $(P^I)^2$ implies 
$I$ is a minimal model of $(P^I)^1$.   
       
$(P^I)^1$ is $P$ with those rules removed whose bodies are 
no satisfied by $I$. Assume, on the contrary, that some
$M\subset I$ is a model of $P$. Since  
$I$ is a minimal model of $(P^I)^1$,
$(P^I)^1$ is not satisfied by $M$.
Since $(P^I)^1\subseteq P$,
$P$ is not satisfied by $M$, a contradiction.
As a result, $I$ is a minimal model of $P$.
This concludes the proof. 
\hfill \qed\\[.1in]
\indent The following lemma is required for the proof of 
Theorem \ref{th-smodel-2}.

\begin{lemma}
\label{lem-smodel-2}
Let $P$ be a positive logic program with ordinary atoms
and $A$ be a literal in $P$.
Let $P'$ be $P$ with each occurrence of $A$ in rule bodies
replaced by a special atom $\theta_A$, and 
each occurrence of $A$ in rule heads
replaced by a special atom $\beta_A$, where
$\theta_A$ is defined in $P'$ by a rule
$\theta_A\leftarrow A$,
and $\beta_A$ is defined in $P'$ by  
two rules $A \leftarrow \beta_A$ and $\beta_A \leftarrow A$.
An interpretation $I$ is a stable model of $P$ if and only if 
$M$ is a stable model of $P'$ with $I = M\setminus  \{\theta_A,\beta_A\}$.
\end{lemma}

\noindent {\bf Proof:} 
Since $\theta_A$ is used only to replace 
$A$ in rule bodies, it can be derived from $P'$
only by applying the rule
$\theta_A\leftarrow A$. That is, if $\theta_A$ 
is in a stable model of $P'$, $A$ must be in the model.
The converse also holds. Therefore, replacing 
$\theta_A$ with $A$ does not change the semantics of $P'$.

For $\beta_A$, the two rules $A \leftarrow \beta_A$ and
$\beta_A \leftarrow A$ express $A \equiv \beta_A$.
Thus, replacing $\beta_A$ with $A$ does not change the semantics of $P'$.

After the above replacement, we transform $P'$ to $P$.
Therefore, $P'$ and $P$ have the same stable models.
\hfill \qed\\[.1in]
\noindent {\bf Proof of Theorem \ref{th-smodel-2}:} 
Let $not\ A$ be a negative literal in the body of a rule $r$
of $P$, which is replaced in $P'$ by a c-atom $A' = (\{A\}, \{\emptyset\})$. 
When $I\not\models not\ A$ (i.e., $A\in I$), we have $I\not\models A'$;
when $I\models not\ A$ (i.e., $A\not\in I$), we have $I\models A'$.
For the former case, $r$ will be removed in the first operation,
from $P$ under the standard Gelfond-Lifschitz transformation,
and from $P'$ under the generalized Gelfond-Lifschitz transformation.
For the latter case, $not\ A$ will be removed from $r$
under the standard Gelfond-Lifschitz transformation,
while $A'$ will be replaced, 
under the generalized Gelfond-Lifschitz transformation,
by a special atom $\theta_{A'}$, where $\theta_{A'}$
is defined by a bodiless rule $\theta_{A'}$ in $P'$. 
In this case, $\theta_{A'}$ can be removed from $P'$.
Let $P^I$ be the standard Gelfond-Lifschitz transformation of $P$ w.r.t. $I$.
We can further remove all rules from $P^I$ whose body contains a positive 
literal $A\not\in I$, since if $I$ is a stable model, $A$
will not be derived from $P^I$ and thus these rules will
not be applicable. These rules will also be removed from $P'$
in the first operation of the generalized Gelfond-Lifschitz transformation, as
$A\not\in I$ implies $I\not\models (\{A\}, \{\{A\}\})$.
As a result, the resulting standard transformation $P^I$ of $P$ is the same as
${P'}^I$ obtained by applying to $P'$ the first two operations
of the generalized Gelfond-Lifschitz transformation,
except that each atom $A$ in $P^I$ is replaced in ${P'}^I$ 
by a c-atom $(\{A\}, \{\{A\}\})$.
Then, after applying to ${P'}^I$ the third and fourth operations
of the generalized Gelfond-Lifschitz transformation,
${P'}^I$ becomes $P^I$ except that for each literal $A$ in $P^I$,
each occurrence of $A$ in rule bodies are
replaced by a special atom $\theta_A$, and 
each occurrence of $A$ in rule heads
replaced by a special atom $\beta_A$, where
$\theta_A$ is defined in ${P'}^I$ by a rule
$\theta_A\leftarrow A$,
and $\beta_A$ is defined in ${P'}^I$ by  
two rules $A \leftarrow \beta_A$ and $\beta_A \leftarrow A$.
By Lemma \ref{lem-smodel-2}, $I$ is a stable model of $P^I$ if and only if 
$M$ is a stable model of ${P'}^I$ with $I = M\setminus  \Gamma$. 
This means that $I$ is a stable model of $P$ if and only if 
it is a stable model of $P'$. \hfill \qed \\[.1in]
\noindent {\bf Proof of Theorem \ref{th-complexity}:} 
The first part of the theorem is straightforward,
as all satisfiable sets of $A$ w.r.t. $T_A^I$
can be obtained simply by comparing each $W\uplus V$ in $A_c^*$
with $T_A^I$ to see if it covers $T_A^I$. 

For the second part of the theorem, the time complexity of 
the generalized Gelfond-Lifschitz transformation
consists of the following three parts: (i) The time complexity
of the standard Gelfond-Lifschitz transformation of $P$
with all c-atoms ignored. This is linear
in the number $|P|$ of rules in $P$. 
(ii) The time complexity of computing all satisfiable sets
of all $n$ c-atoms. As just proved above, it is bounded by $O(n*M_{A_c^*})$.
(iii) The time complexity of introducing new rules for all $n$ c-atoms.
Assume that it takes constant time to introduce a new rule
for a special atom $\theta_A$ or $\beta_A$
(see the third and fourth operations).
Then, the time complexity of this part is bounded by $O(n*(M_{A_c^*} + M_{A_d} + 1))$,
as the generalized Gelfond-Lifschitz transformation
introduces at most $2*n$ special atoms 
(one $\theta_A$ and one $\beta_A$ for each c-atom $A$), each accompanied by
at most $M_{A_c^*}$ (for $\theta_A$) or $M_{A_d}+1$ (for $\beta_A$) new rules. 
The total time complexity of the generalized 
Gelfond-Lifschitz transformation is then bounded by
$O(|P| + n* (2M_{A_c^*} + M_{A_d} + 1))$. \hfill \qed \\[.1in]
\noindent {\bf Proof of Theorem \ref{th-complexity-2}:} 
For normal constraint programs, since our stable model semantics 
coincides with
that of Son et al. \cite{SPT07}, the complexity of the latter semantics applies, 
which is 
known to be NP-complete (stated in \cite{LPST07} as part of computation-based semantics and  proved in \cite{YYLS07}).

It is known that the decision problem for disjunctive programs 
(without c-atoms) is $\Sigma_P^2$-complete \cite{disjunctive-complexity}.
Since disjunctive programs are 
disjunctive constraint programs, the decision problem is at least as 
hard as for disjunctive programs, i.e., it is $\Sigma_P^2$-hard. To see that 
the problem is in $\Sigma_P^2$, we first note that replacing 
c-atoms by their 
abstract
representations takes polynomial time, in the size of $P$ 
(c.f. Theorem \ref{construction-abstract-rep}), so does 
the generalized Gelfond-Lifschitz transformation (c.f. 
Theorem \ref{th-complexity})
for a given interpretation $I$. 
Then, to determine whether $M$ is a minimal model of 
the generalized Gelfond-Lifschitz transformation $P^I$ (cf. Definition 
\ref{my-stablemodel})
is to determine 
whether $M$ is a minimal model of a positive disjunctive program. Therefore, 
the fact that the latter is in 
$\Sigma_P^2$ implies that the former is also in 
$\Sigma_P^2$.
\hfill \qed \\[.1in]
\noindent {\bf Proof of Theorem \ref{sat-set}:} 
($\Longrightarrow$) Assume $R\models_I A$. By Definition \ref{cond-sa}, 
$R\models A$ and for every $S'$ such that $R\cap A_d\subseteq S'$
and $S'\subseteq T_A^I$, we have $S'\in A_c$.
By Definition \ref{ab-1},  
the collection of abstract $(R\cap A_d)$-prefixed power sets of $A$
contains $R\cap A_d\uplus S_i$ with $S_i\supseteq T_A^I \setminus  (R\cap A_d)$,
which covers all $S'$ with $R\cap A_d\subseteq S'\subseteq T_A^I$. 
By Definition \ref{ab-2}, $A_c^*$ contains an 
abstract prefixed power set $W\uplus V$ such that
$R\cap A_d\uplus S_i$ is included in $W\uplus V$.
Since $S_i\supseteq T_A^I \setminus  (R\cap A_d)$,
$R\cap A_d\uplus T_A^I \setminus  (R\cap A_d)$ 
is included in $R\cap A_d\uplus S_i$,
hence $R\cap A_d\uplus T_A^I \setminus  (R\cap A_d)$ 
is included in $W\uplus V$. Note that in this case,
$W\subseteq R\cap A_d$, and since $W\uplus V$ covers $T_A^I$, 
$W$ is a satisfiable set of
$A$ w.r.t. $T_A^I$.  

($\Longleftarrow$) Assume that $A_c^*$ has an 
abstract prefixed power set $W\uplus V$ 
such that $R\cap A_d\uplus T_A^I \setminus  (R\cap A_d)$ 
is included in $W\uplus V$. Then, $W\uplus V$ covers the 
whole collection covered by  
$R\cap A_d\uplus T_A^I \setminus  (R\cap A_d)$.
This means that $W\uplus V$ covers every 
$S'$ with $R\cap A_d\subseteq S'$
and $S'\subseteq T_A^I$. Since $W\uplus V$ is in $A_c^*$,
this collection covered by $W\uplus V$ is included in $A_c$
and thus $R\models A$. By Definition \ref{cond-sa}, 
we have $R\models_I A$. Note again that in this case,
$W\subseteq R\cap A_d$ and $W$ is a satisfiable set of
$A$ w.r.t. $T_A^I$. 
\hfill \qed \\[.1in]
\noindent {\bf Proof of Theorem \ref{equal-models}:} 
Let $P^I$ be the generalized Gelfond-Lifschitz transformation.
Since $P^I$ is a positive normal logic program, it has a least 
model which is the fixpoint $T_{P^I}^\infty(\emptyset)$ with
$T_{P^I}^0(\emptyset) = \emptyset$ and $T_{P^I}^{i+1}(\emptyset) = T_{P^I}(T_{P^I}^i(\emptyset))$,
where the operator $T_{P^I}$ is defined by
\[T_{P^I}(R) = \left\{A\left| \begin{array}{l}
                     \exists r\in {P^I}: R\models body(r),\\
                     head(r) = A
                    \end{array}
                      \right. \right\}\]
We want to prove, by induction on $i\geq 0$, 
that $T_P^i(\emptyset, I) = T_{P^I}^{3i}(\emptyset)  \setminus   \Gamma$. 
As induction basis, when $i=0$, $T_P^0(\emptyset, I) = T_{P^I}^{3*0}(\emptyset) = \emptyset$. 
For induction hypothesis, assume that for any $i\leq k$ we have 
$T_P^i(\emptyset, I) = T_{P^I}^{3i}(\emptyset) \setminus   \Gamma$. 
Now consider $i = k+1$. 

($\Longrightarrow$) Assume that $I$ is a stable model under Son et al.'s fixpoint definition.
We first prove that for each atom $B$ derived in $T_P^{k+1}(\emptyset, I)$
(i.e., $B\in T_P^{k+1}(\emptyset, I)$ but $B\not\in T_P^k(\emptyset, I)$),
we have $B\in T_{P^I}^{3(k+1)}(\emptyset)$. 
%That is, we want to prove $I\subseteq T_{P^I}^\infty(\emptyset)$. 
By Definition \ref{def-spt06}, there is a rule $r$ in $P$ 
of the form 
\[r:\quad (\{B\}, \{\{B\}\})\leftarrow A_1, ..., A_m\] 
such that $T_P^k(\emptyset, I)\models_I body(r)$.
Consider an arbitrary c-atom $A_j$ in $body(r)$. Note that    
$T_P^k(\emptyset, I)\models_I A_j$. By Theorem \ref{sat-set},
there is a satisfiable set $W$ of
$A_j$ w.r.t. $T_{A_j}^I = I\cap A_{j_d}$ such that 
$W\subseteq T_P^k(\emptyset, I)\cap A_{j_d}$. 
Let $W = \{D_1, ..., D_t\}\subseteq T_P^k(\emptyset, I)$.
The generalized Gelfond-Lifschitz transformation
$P^I$ must contain the following rules:
\begin{tabbing} 
$\qquad (1)\  \beta_B\leftarrow \theta_{A_1},...,\theta_{A_m},$  \\ 
$\qquad (2)\  B \leftarrow \beta_B,$  \\ 
$\qquad (3)\  \theta_{A_j}$ \= $\leftarrow D_1, ...,D_t.$ 
%$\ $ for each $A_j$ in $body(r)$ and
%each\\
%\>  satisfiable set $\{D_1, ..., D_t\}$ of $A_j$ w.r.t. $T_{A_j}^I$  
\end{tabbing}
By the induction hypothesis, $\{D_1, ..., D_t\}\subseteq T_{P^I}^{3k}(\emptyset)$.
Due to this, rule (3) can be applied, leading to $\theta_{A_j}\in T_{P^I}^{3k+1}(\emptyset)$.
This process applies to all c-atoms $A_j$ in $body(r)$ so that 
$\theta_{A_1},...,\theta_{A_m}$ are all in $T_{P^I}^{3k+1}(\emptyset)$.
Rule (1) is then applied, leading to $\beta_B\in T_{P^I}^{3k+2}(\emptyset)$.
Then, rule (2) is applied, leading to $B\in T_{P^I}^{3k+3}(\emptyset)$.

The above induction shows that for any atom $B\in T_P^i(\emptyset, I)$, we have 
$B\in T_{P^I}^{3i}(\emptyset)$. When $i\rightarrow \infty$,
$T_P^\infty(\emptyset, I) \subseteq T_{P^I}^\infty(\emptyset)$.
Since $I$ is a stable model under Son et al.'s fixpoint definition
with $T_P^\infty(\emptyset, I) = I$ 
and contains no special atoms, we have
$I\subseteq T_{P^I}^\infty(\emptyset)  \setminus   \Gamma$. 

Next, we prove that when $I$ is a stable model under Son et al.'s fixpoint definition,
we have $T_{P^I}^\infty(\emptyset) \setminus   \Gamma \subseteq I$. 
For any (non-special) atom $B$ derived in $T_{P^I}^{3(k+1)}(\emptyset)$,
there must be a rule $r$ as above in $P$ and
a rule of form (2) in $P^I$ derived from $r$
such that $\beta_B$ is derived in $T_{P^I}^{3k+2}(\emptyset)$ by applying rule (1)
where each $\theta_{A_j}$ is satisfiable in $T_{P^I}^{3k+1}(\emptyset)$
and at least one $\theta_{A_j}$ is derived 
in $T_{P^I}^{3k+1}(\emptyset)$ by applying rule (3)
where each atom $D_j$ is satisfiable in $T_{P^I}^{3k}(\emptyset)$. 
By the induction hypothesis, $T_P^k(\emptyset, I) = T_{P^I}^{3k}(\emptyset)  \setminus   \Gamma$,
so $T_P^k(\emptyset, I)\models \{D_1, ..., D_t\}$. 
Let $W = \{D_1, ..., D_t\}$. Since $W$ comes from rule (3), it  
is a satisfiable set of $A_j$ w.r.t. $T_{A_j}^I = I\cap A_{j_d}$.
By Definition \ref{sat-sets},
$A_{j_c}^*$ contains an abstract $W$-prefixed power set $W\uplus V$
covering $T_{A_j}^I$. So, $W\uplus T_{A_j}^I \setminus  W$ is included in $W\uplus V$.
Since $T_P^k(\emptyset, I)\subseteq T_P^\infty(\emptyset, I) = I$, we have 
$W\subseteq T_P^k(\emptyset, I) \cap A_{j_d} \subseteq T_{A_j}^I$.
By Theorem \ref{t1}, 
$T_P^k(\emptyset, I) \cap A_{j_d} \uplus T_{A_j}^I \setminus  (T_P^k(\emptyset, I) \cap A_{j_d})$ 
is included in $W\uplus T_{A_j}^I \setminus  W$, thus it
is included in $W\uplus V$.
By Theorem \ref{sat-set}, $T_P^k(\emptyset, I) \models_I A_j$.
This holds for all $A_j$ in $body(r)$. By Definition \ref{def-spt06},
$B$ is in $T_P^{k+1}(\emptyset, I)$.
This induction shows that for any non-special atom $B\in T_{P^I}^{3i}(\emptyset)$,
we have $B\in T_P^i(\emptyset, I)$. When $i\rightarrow \infty$,
$T_{P^I}^\infty(\emptyset)  \setminus   \Gamma\subseteq T_P^\infty(\emptyset, I)=I$.

The above proof concludes that when $I$ is a stable 
model under Son et al.'s fixpoint definition,
$T_{P^I}^\infty(\emptyset) \setminus   \Gamma = I$. 
Hence, by Definition \ref{my-stablemodel}
$I$ is a stable model derived from the 
generalized Gelfond-Lifschitz transformation.

($\Longleftarrow$) Assume that $I$ is a stable model, with
$T_{P^I}^\infty(\emptyset)  \setminus   \Gamma = I$, derived from the 
generalized Gelfond-Lifschitz transformation. Copying the same proof
as the first part above, we can prove that any non-special atom 
$B$ derived in $T_P^{k+1}(\emptyset, I)$ is in $T_{P^I}^{3(k+1)}(\emptyset)$. 
That is, $T_P^\infty(\emptyset, I)\subseteq T_{P^I}^\infty(\emptyset)  \setminus   \Gamma = I$. 
Next, we prove the converse part: $I \subseteq T_P^\infty(\emptyset, I)$. 

For any (non-special) atom $B$ derived in $T_{P^I}^{3(k+1)}(\emptyset)$,
there must be a rule $r$ as above in $P$ and
a rule of form (2) in $P^I$ derived from $r$
such that $\beta_B$ is derived in $T_{P^I}^{3k+2}(\emptyset)$ by applying rule (1)
where each $\theta_{A_j}$ is satisfiable in $T_{P^I}^{3k+1}(\emptyset)$
and at least one $\theta_{A_j}$ is derived 
in $T_{P^I}^{3k+1}(\emptyset)$ by applying rule (3)
where each atom $D_j$ is satisfiable in $T_{P^I}^{3k}(\emptyset)$. 
By the induction hypothesis, $T_P^k(\emptyset, I) = T_{P^I}^{3k}(\emptyset)  \setminus   \Gamma$,
so $T_P^k(\emptyset, I)\models \{D_1, ..., D_t\}$. 
Let $W = \{D_1, ..., D_t\}$. Since $W$ comes from rule (3), it
is a satisfiable set of $A_j$ w.r.t. $T_{A_j}^I = I\cap A_{j_d}$.
By Definition \ref{sat-sets},
$A_{j_c}^*$ contains an abstract $W$-prefixed power set $W\uplus V$
covering $T_{A_j}^I$. So, $W\uplus T_{A_j}^I \setminus  W$ is included in $W\uplus V$.
Note that $T_P^k(\emptyset, I)\subseteq I$ because 
$T_{P^I}^{3k}(\emptyset)  \setminus   \Gamma \subseteq T_{P^I}^\infty(\emptyset)  \setminus   \Gamma = I$. 
Then, we have 
$W\subseteq T_P^k(\emptyset, I) \cap A_{j_d} \subseteq T_{A_j}^I$.
By Theorem \ref{t1}, 
$T_P^k(\emptyset, I) \cap A_{j_d} \uplus T_{A_j}^I \setminus  (T_P^k(\emptyset, I) \cap A_{j_d})$ 
is included in $W\uplus T_{A_j}^I \setminus  W$, thus it
is included in $W\uplus V$.
By Theorem \ref{sat-set}, $T_P^k(\emptyset, I) \models_I A_j$.
This holds for all $A_j$ in $body(r)$. By Definition \ref{def-spt06},
$B$ is in $T_P^{k+1}(\emptyset, I)$.
This induction shows that for any non-special atom $B\in T_{P^I}^{3i}(\emptyset)$,
we have $B\in T_P^i(\emptyset, I)$. When $i\rightarrow \infty$,
$T_{P^I}^\infty(\emptyset)  \setminus   \Gamma \subseteq T_P^\infty(\emptyset, I)$.
That is, $I \subseteq T_P^\infty(\emptyset, I)$. 

The above proof concludes that when $I$ is a stable 
model derived from the generalized Gelfond-Lifschitz transformation,
$T_P^\infty(\emptyset, I) = I$. 
Hence, $I$ is also a stable model under Son et al.'s fixpoint definition. 
\hfill \qed \\[.1in]
\noindent {\bf Proof of Lemma \ref{positive-prog}:} 
First we note that, under the assumptions of basic programs in this section,
part 4 in Definition \ref{my-stablemodel} can be omitted. Thus, 
that $I$ is a stable model of $P$ if and only if 
$M = I \cup \Gamma_\theta$ 
is the least model of the generalized Gelfond-Lifschitz transformation $P^I$,
if and only if $M$ is 
the least model of the standard Gelfond-Lifschitz transformation $P_n^M$.
\hfill \qed \\[.1in]
\noindent {\bf Proof of Theorem \ref{dependency-theorem}:} 
We know that the same claims hold for normal 
programs ((1), (3)  and (4) are due to \cite{fages94}, and
(2) due to \cite{y-y-94}), where the dependency graph is defined as: 
for each rule 
$a \leftarrow b_1, ..., b_m, not\ c_1, ..., not \ c_n$
in a normal program, there
is a positive edge from $a$ to each $b_i$, $1 \le i \le m$, and 
a negative edge from $a$ to each $c_j$, $1 \le j \le n$. 
Let us denote
by $G^N_P$ the dependency graph for a normal program $P$.
Recall that we use $G_P$ to denote the dependency graph for a basic program $P$.

Let $P$ be a basic program and $P_n$ be its normal program translation.
By definition, for any positive edge $u \rightarrow^+ v$ in $G_P$, 
there is a path 
$u \rightarrow^+ \theta_{A_i} \rightarrow^+ v$ in $G^N_{P_n}$,
for some new symbol $\theta_{A_i}$,
and vice versa. Similarly, 
for any negative edge $u \rightarrow^- v$ in $G_P$, 
there is a path $u \rightarrow^+ \theta_{A_i} \rightarrow^- v$ in $G^N_{P_n}$,
and vice versa. Therefore, for any loop $L$ in $G_P$, there is a loop
$L'$ in $G^N_{P_n}$
with some additional positive edges to new symbols,
and vice versa. Therefore, there is a one-to-one correspondence between loops 
in $G_P$ and those in $G^N_{P_n}$, modulo the new symbols $\theta_{A_i}$. 

Notice that the extra positive edges have no effect on the type of 
the loops based on negative 
dependency; i.e., for any odd cycle in $G_P$, the same  
odd cycle with some additional positive edges is
in $G^N_{P_n}$, and vice versa; similarly for even cycles.

Let $P$ be a basic program.
Suppose $P$ is call-consistent, i.e.,
$P$ has no odd cycles in $G_P$.
By the one-to-one correspondence between cycles, 
$P_n$ has no odd cycles in $G^N_{P_n}$.
Thus, according to \cite{fages94}, 
a stable model, say $M$, exists for $P_n$. 
By Lemma \ref{positive-prog},
$M_{|At(P)}$ is a stable model of $P$. This proves claim (1).
Now assume $P$ has more than one stable model, say $M_1$ and $M_2$ 
(and possibly others). By Lemma \ref{positive-prog}, $P_n$ has stable models
$S_1$ and 
$S_2$ such that  
${S_1}_{|At(P)} = M_1$ and ${S_2}_{|At(P)} = M_2$. Thus, according to 
\cite{y-y-94}, $P_n$ has an even loop in $G^N_{P_n}$, and it follows
that 
$P$ has an even loop in $G_P$. 
This proves claim (2). Now assume 
$P$ is acyclic in $G_P$.
Then $P_n$ is acyclic in $G^N_{P_n}$.
By Lemma \ref{positive-prog} again,
that $P_n$ has a unique 
stable model implies the same for $P$. This shows claim (3). 
Finally, suppose $P$ has no positive cycles in $G_P$.
Let $M$ be a supported model of $P$. 
We can extend 
$M$ to be a supported model of $P_n$ by adding 
extra symbols $\delta_{A_i}$ in the following way: whenever a rule of 
the form (\ref{unfolding-positive}) in $P$ supports atom $H$ in $M$, add 
$\theta_{A_i}$ ($1 \le i  \le n$) of the rule (\ref{extra-symbol}) in 
$P_n$
to $M$. Let the resulting set be $S$. That is, $M = S_{|At(P)}$.
Clearly, $S$ is a supported model of $P_n$. Since $P_n$ has no positive cycle in $G^N_{P_n}$, $S$ is a 
stable model of $P_n$, and by Lemma \ref{positive-prog},
$M$ is a stable model of $P$. This proves claim (4).
\hfill \qed  

\bibliographystyle{acmtrans}
\bibliography{tplp3}

\end{document}